\documentclass{article}
\pdfoutput=1
\usepackage{hyperref}
\usepackage{natbib}
\usepackage{graphicx,color}
\usepackage{amsmath}
\usepackage{geometry}
\usepackage{verbatim}
\usepackage{amsthm}
\newtheorem{definition}{Definition}  
\usepackage{statex2}
\usepackage{amsbsy}

\def\C {\,|\:}

\newcommand{\bi}{\begin{itemize}}
\newcommand{\ib}{\end{itemize}}
\newcommand{\p}{\item}
\newcommand{\be}{\begin{enumerate}[(i)]}
\newcommand{\eb}{\end{enumerate}}

\newtheorem{theorem}{Theorem}
\title{
\Large Causal Inference with the Instrumental Variable Approach
and Bayesian Nonparametric Machine Learning
}
\author{%
Robert McCulloch 
\footnote{%
Robert E. McCulloch, School of Mathematical and Statistical Sciences, Arizona State University,
Robert.McCulloch@asu.edu.
Rodney Sparapani, Division of Biostatistics, Medical College of Wisconsin.
Brent Logan, Division of Biostatistics, Medical College of Wisconsin.
Purushottam Laud, Division of Biostatistics, Medical College of Wisconsin.
} \and Rodney Sparapani \and Brent Logan \and Purushottam Laud 
}

\begin{document}

\maketitle
\thispagestyle{empty}

\begin{abstract}
\noindent 

We provide a new flexible framework for inference with the
instrumental variable model.  Rather than using linear specifications,
functions characterizing the effects of instruments and other
explanatory variables are estimated using machine learning via
Bayesian Additive Regression Trees (BART).  Error terms and their
distribution are inferred using Dirichlet Process mixtures.  Simulated
and real examples show that when the true functions are linear, little
is lost.  But when nonlinearities are present, dramatic improvements
are obtained with virtually no manual tuning.

\end{abstract}

\newpage

\setcounter{page}{0}
\tableofcontents
\thispagestyle{empty}

\newpage

\section{Introduction}\label{sec:introduction}

The instrumental variable (IV) approach has long been a cornerstone of
causal inference from both the theoretical and applied perspectives.
For example, the distribution-free method of two-stage least squares
(TSLS) goes back to \cite{Thei53} and it is based on earlier IV work
that goes back decades further such as \cite{Wrig28}.  The focus on
distribution-free methods is paramount since the reliance on
parametric assumptions has been roundly criticized \citep{LaLo86}.
Therefore, there has been a movement towards nonparametric methods
that do not rely on precarious restrictive assumptions such as
functional forms and/or convenient choices of distributions
\citep{AngrImbe95}.  Conversely, unrestricted nonparametric approaches
may have theoretical challenges such as the lack of causal
identification \citep{Pear09}.  While in practical performance,
distribution-free methods such as TSLS have come under attack as more
biased and less powerful than Ordinary Least Squares (OLS) with
standard errors generated by either bootstrapping or jack-knifing
\citep{Youn19}.

We take a Bayesian approach to IV as many others have before us
\citep{ImbeRubi97,RossAlle05,CHMR08,ROSSI14}.  For example,
\cite{RossAlle05} take a Bayesian parametric approach that we will
refer to as {\it linear-normal} or {\it lin-nor} (for linear IV with
normal errors) based upon the linear structural equations of TSLS.
\cite{CHMR08} expand on this previous work via a semi-parametric
method that relaxes the parametric error distribution with Dirichlet
Process Mixtures (DPM) \citep{EW95} while retaining the linear model
structural equations of TSLS: we will refer to this method as {\it
linear-DPM} or {\it lin-DPM}.  For a comprehensive exposition of the
IV framework from the Bayesian perspective, along with Bayesian
nonparametric priors, see \cite{ROSSI14}.

Herein, we propose a new nonparametric method based on Bayesian
Additive Regression Trees (BART) \citep{ChipGeor10} and DPM
capable of handling structural equations that may be non-linear and/or
may have non-normal errors.  We will refer to our new method as {\it
IVBART} which we describe in Section~\ref{flex-mod}.  In
Section~\ref{simulated}, we explore our new method with simulated data
sets and compare with lin-nor and lin-DPM.  Section~\ref{card} is
where we delve into a real data set that
demonstrates our new method to estimate the monetary returns of
post-secondary education (as have others \citep{Card93,CHMR08}).  In
Sections~\ref{gibbs} and \ref{details},
we provide the details for implementing our new method via Markov
chain Monte Carlo (MCMC) sampling of the posterior.
Section~\ref{sec:conclusion} concludes the article with a brief
discussion of the merits of our new method and some potential future
directions for extensions.  In the Appendix, we provide a brief
introduction to the {\bf ivbart R} package that implements our new
method and a proof of the causal identification of our new method.

\section{Flexible IV Modeling}\label{flex-mod}

In this section we present our model and basic computational approach.

The classic TSLS linear approach to IV modeling is expressed by 
the following two equations:

\begin{eqnarray}\label{liniv}
T_i & = & \mu_T + \gamma' z_i + \alpha' x_i + \epsilon_{Ti} \label{liniv1} \\
Y_i & = & \mu_Y + \beta \, T_i + \delta' x_i +
          \epsilon_{Yi} \label{liniv2}\ .
\end{eqnarray}

Equation \ref{liniv1} is the {\it treatment} or {\it first stage} equation
where $z$ are the instruments and $x$ are the confounders.
Equation \ref{liniv2} is the {\it outcome} or {\it second stage} equation.
We do not want to assume that the errors $\epsilon_{Ti}$ and $\epsilon_{Yi}$ are independent
since unmeasured variables may be affecting both $T$ and $Y$.
The idea of the model is that the instrumental variable $z$ provides a
{\it source of variation} in $T$, such as a {\it natural experiment}, that
is analogous to the variation induced by an experimenter who
controls the value of $T$ assigned.

Our goal is to eliminate the need to assume that the relationships are
linear and to make minimal assumptions about the nature of the errors.
We simply replace the linear functions in Equations \ref{liniv1} and
\ref{liniv2} above with general functions.  To facilitate the modeling
of the errors, we combine the error terms with the means as follows:

\begin{eqnarray}\label{nliniv}
T_i & = & f(z_i,x_i) + \epsilon_{Ti} \label{n-liniv1} \\
Y_i & = & \beta \, T_i + h(x_i) + \epsilon_{Yi}\ . \label{n-liniv2}
\end{eqnarray}

We model each of the functions $f$ and $h$ using the BART methodology and we model the errors using
Bayesian nonparametrics as in \cite{CHMR08}.
Our hope is that the model given by Equations
\ref{n-liniv1} and \ref{n-liniv2} will be tremendously appealing to
applied investigators.  Our belief is that the relaxation of the linearity
assumption is a much more powerful elaboration of the model than the
relaxation of the normal error assumption.  In practice, applied
investigations typically struggle to deal with potential nonlinearity by
transformations of $z$ and/or $x$ in the model.  This leads to an
unappealing model specification since $z$ and/or $x$ have been
transformed from their natural representation.  A basic goal of
Machine Learning is to learn the functions $f$ and $h$ fairly
automatically from the data.  In practice, Machine Learning can
involve a complex model training phase and extensive use of
cross-validation to select the tuning parameters.  Our choice of BART
as the method for learning has some fundamental advantages. 

\bi 
\p BART is able to learn high-dimensional, complex, non-linear
relationships \p BART is a fully Bayesian procedure with an effective
MCMC algorithm that inherently provides an assessment of uncertainty.
\p BART often obtains an adequate fit with minimal tuning. \p Multiple
additive BART models can be embedded in a larger model \\ 
(as in Equations \ref{n-liniv1} and \ref{n-liniv2} above).
\ib

%

To model the error terms we use the Dirichlet process mixture (DPM) approach 
of \cite{EW95}.
A simple way to think about the DPM model is to let
$$
\epsilon_i = (\epsilon_{Ti},\epsilon_{Yi})' \sim N(\mu_i,\Sigma_i)
$$
so that 
each error $\epsilon_i$  has its own mean $\mu_i$ and variance matrix $\Sigma_i$.
Of course, this model
is too flexible without further structure.
Let $\theta_i = (\mu_i,\Sigma_i)$.
The DPM method adds a hierarchical model for the set of  $\theta_i$ so that
there is a random number of unique values.
Each observation can have its own $\theta$, but observations share $\theta$ values
so that the number of unique values is far less than the sample size.
This reduces the effective complexity of the parameter space.

The DPM hierarchical model draws a discrete distribution using the  Dirichlet process (DP)
and then draws the $\theta_i$ from the discrete distribution.  Because the distribution
is discrete, with positive probability, some of the $\theta_i$ values will be repeats.
To simplify notation, let $\{x_i\}$ represent $\{x_i\}_{i=1}^n$ in (\ref{eq:DPM}).
Letting $G$ denote the random discrete distribution, our hierarchical model is:
\begin{equation}\label{eq:DPM}
\{\epsilon_i\} \C \{\theta_i\}, \;\;  \{\theta_i\}\C G, \;\;  G \C G_0,\alpha
\end{equation}
where
$$
\epsilon_i \sim N(\mu_i,\Sigma_i), \;\; \theta_i = (\mu_i,\Sigma_i) \sim G, \;\; G \sim DP(G_0,\alpha).
$$

$DP$ denotes the Dirichlet process distribution over discrete
distributions given parameters $G_0$ and $\alpha$.  We refer the
reader to \cite{CHMR08} for the complete details.  Briefly, to
motivate our prior choices, we need some basic intuition about how the
choices for $G_0$ and $\alpha$ affect the inference.  $G_0$ is the
{\it central} distribution over the space of $\theta=(\mu,\Sigma)$.
The atoms of $G$ are independent and identically distributed, 
or {\it iid}, draws from $G_0$.  The {\it concentration}
parameter $\alpha$ determines the distribution of the weights given to
each atom of the discrete $G$.  A larger $\alpha$ tends to give you a
discrete $G$ with more atoms receiving non-negligible weight.  A
smaller $\alpha$ means only of few of the weights are likely to be
large so that $G$ tends to have most of its mass concentrated on just
a few atoms.  In terms of the {\it mixture of normals} interpretation,
$G_0$ tells us what normal distributions are likely (what
$\theta = (\mu,\Sigma)$ are likely); and $\alpha$ tells us how many
normals there are and with what weight.

Thus, our parameter space can be thought as:
$$
f,\; h,\; \beta,\; \{\theta_i\}.
$$

Our computational algorithm is the obvious Gibbs sampler
\citep{GelfSmit90}:

\begin{eqnarray}\label{gibbs}
f \C h,\beta,\{\theta_i\},D \label{gibbsf} \\
h \C f,\beta,\{\theta_i\},D \label{gibbsh} \\
\beta \C f,h,\{\theta_i\},D \label{gibbsb} \\
\{\theta_i\} \C f,h,\beta,D \label{gibbsb} \\
\end{eqnarray}
where $D$ denotes the observed data $\{T_i,Y_i,x_i,z_i\}_{i=1}^n$.
Most of these draws are straightforward and follow \cite{CHMR08}.
The exception is the draw of $f$ where the nonlinearity calls for special treatment.
Details of the draws are given in Section~\ref{gibbs}.

\section{Simulated Examples}\label{simulated}

In this section we illustrate our methodology on simulated data.
The parameters of our model we must choose in order to simulate data are the value of $\beta$,
the $(f,h)$ pair of functions, and the error distribution.
We must also choose distributions to draw $x$ and $z$ from and the sample size.
We will always use $\beta=1$.


We will consider a nonlinear pair of $(f,h)$:

\begin{eqnarray}\label{eq:sim-nonlinfuns}
f(x,z) & = & x_1 + .5 \, x_1 x_2 + .5 x_2^2 + z_1 + z_2 x_1 + .5 z_2^2  \label{eq:sim-nonlinfuns-f} \\
h(x) & = & x_1 - .25 x_1 x_2^3 + x_3 \label{eq:sim-nonlinfuns-h}
\end{eqnarray}

and a linear pair of $(f,h)$:

\begin{eqnarray}\label{eq:sim-linfuns}
f(x,z) & = & x_1 + x_2 + x_3 + z_1 + z_2 \label{eq:sim-linfuns-f}\\
h(x) & = & x_1 - x_2 + .5 x_4\ . \label{eq:sim-linfuns-h}
\end{eqnarray}

The nonlinear functions are chosen to be simple polynomials.
This is not too different from what a practitioner might try but 
a practitioner may have difficulty finding the exact right polynomial terms in practice.
The linear functions are chose to be a simple as possible while having $h$ not too similar to the $x$ part of $f$.

For the error distribution we use:
\begin{eqnarray}\label{eq:sim-error}
\epsilon_T & = & \sigma_T \, Z_T \\
\epsilon_Y & = & \gamma Z_T + \sigma_Y Z_Y
\end{eqnarray}
where $(Z_T,Z_Y)$ are indepenent $t_\nu$ random variables with $\nu=5$.
We let $\sigma_T = 1$ and $(\gamma,\sigma_Y) = (\frac{1}{\sqrt{2}},\frac{1}{\sqrt{2}})$. 
Note that a linear combination of independent $t$ random variables is not a $t$ random variable so that the error $\epsilon_Y$ has
a non-standard distribution.
Clearly $\gamma$ controls the degree of dependence between $\epsilon_T$ and $\epsilon_Y$.
With these choices, both errors have the same variance as the $Z$'s and the correlation is $1/\sqrt{2} \approx .707$.
The pair of errors $(\epsilon_{Ti},\epsilon_{Yi})$ are iid over observations.

Each coordinate of both $x$ and $z$ are iid uniform on the interval $(-2,2)$.
For $x$, we simulate $x_j, j=1,2,\ldots,10$ and for $z$ we simulated $z_j, j=1,2, \ldots 5$.
So, there are 10 $x$ variables and 5 potential $z$ instruments.
Notice that in the nonlinear case (Equations  \ref{eq:sim-nonlinfuns-f} and \ref{eq:sim-nonlinfuns-h}),
$f$ uses only $(x_1,x_2,z_1,z_2)$ while $h$ uses only $(x_1,x_2,x_3)$.
The method is given all 10 $x$ and all 5 $z$ and the two BART models in IVBART have to learn which variables matter.
In the linear case (Equations \ref{eq:sim-linfuns-f} and \ref{eq:sim-linfuns-h}),
$f$ and $h$ use $(x_1,x_2,x_3,z_1,z_2)$ and $(x_1,x_2,x_4)$ respectively.

We consider four different simulation scenarios by letting the sample
size $n$ be 2,000 or 500 and letting the functions be nonlinear or
linear.  We draw 90 samples and run MCMC estimation of each of the
three models IVBART, linear-normal, and linear-DPM on each of the 90
samples.

All IVBART results are obtained using a default prior specification
explained in Section~\ref{subsec:sim-prisens} and, in more detail, in
Section~\ref{details}.  Results for the linear-normal and linear-DPM
models are obtained using the default prior specifications provided by
the functions \texttt{rivGibbs} and \texttt{rivDP} in the \texttt{R}
package \texttt{bayesm} \citep{RPbayesm}.

\subsection{Inference for $\beta$}\label{subsec:sim-beta-inf}

\begin{figure}
\hspace*{-.3in}
\includegraphics[scale=.3]{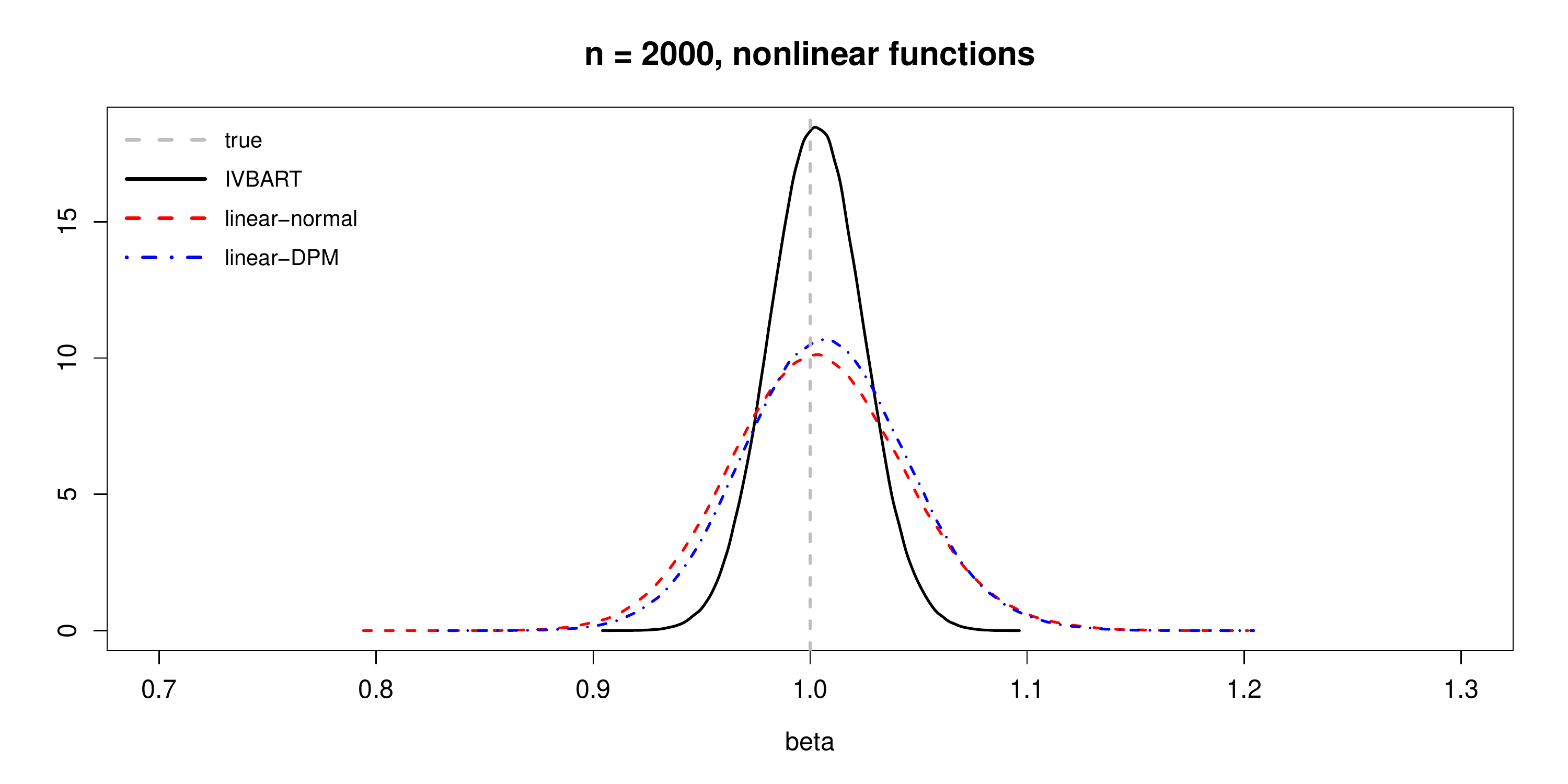}
\hspace{.1in}
\includegraphics[scale=.3]{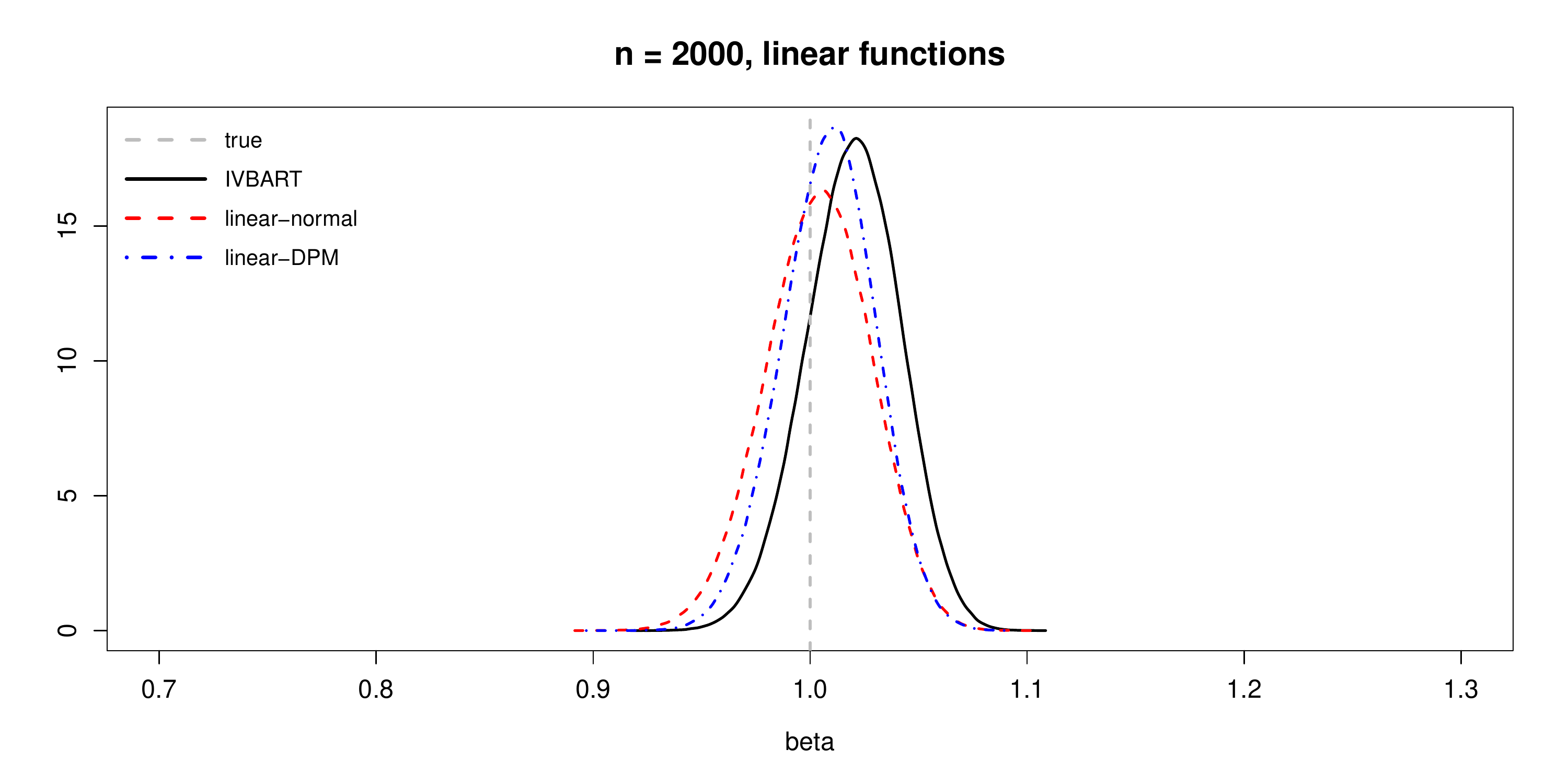} \\
\hspace*{-.3in}
\includegraphics[scale=.3]{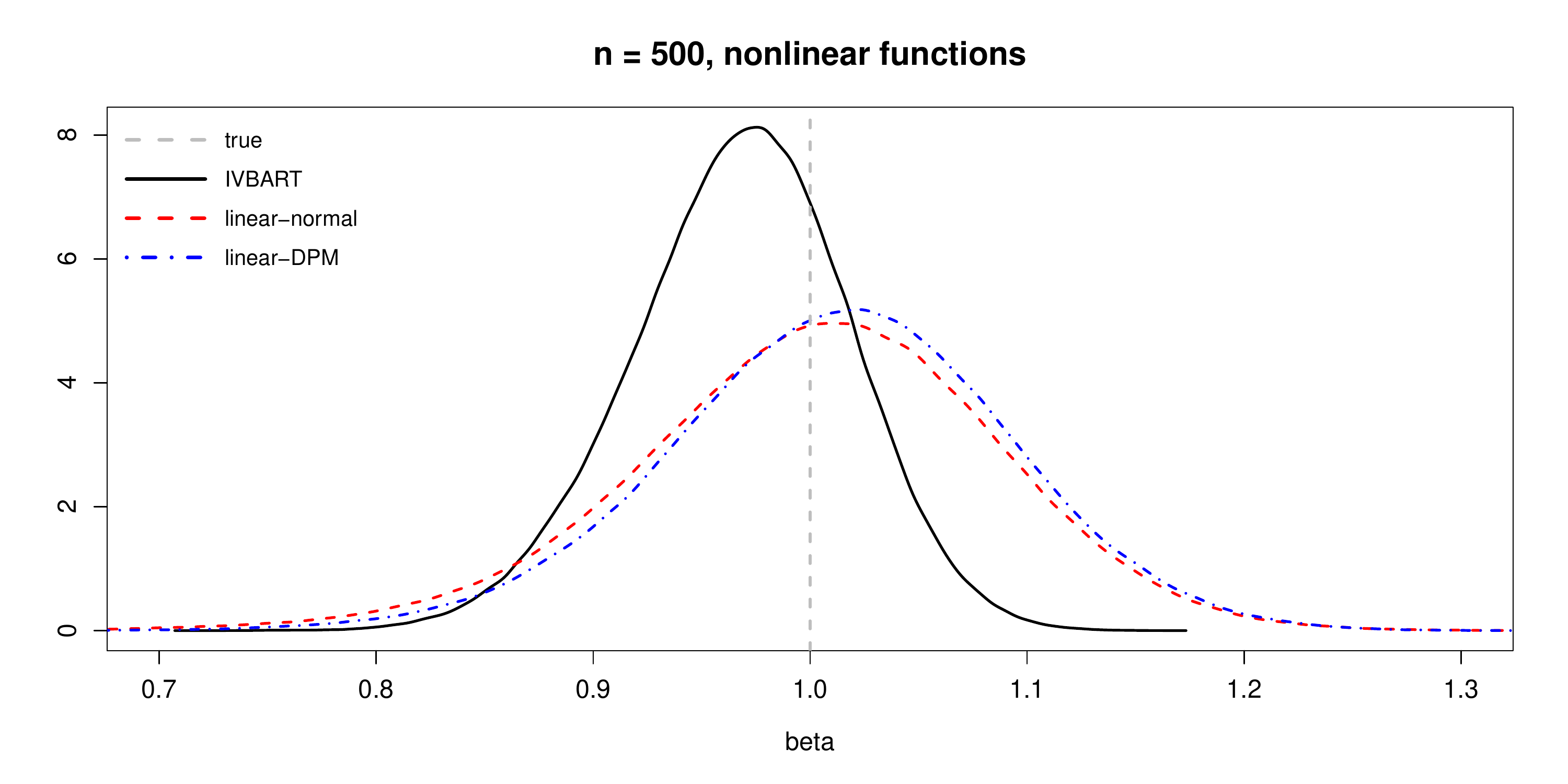}
\hspace{.1in}
\includegraphics[scale=.3]{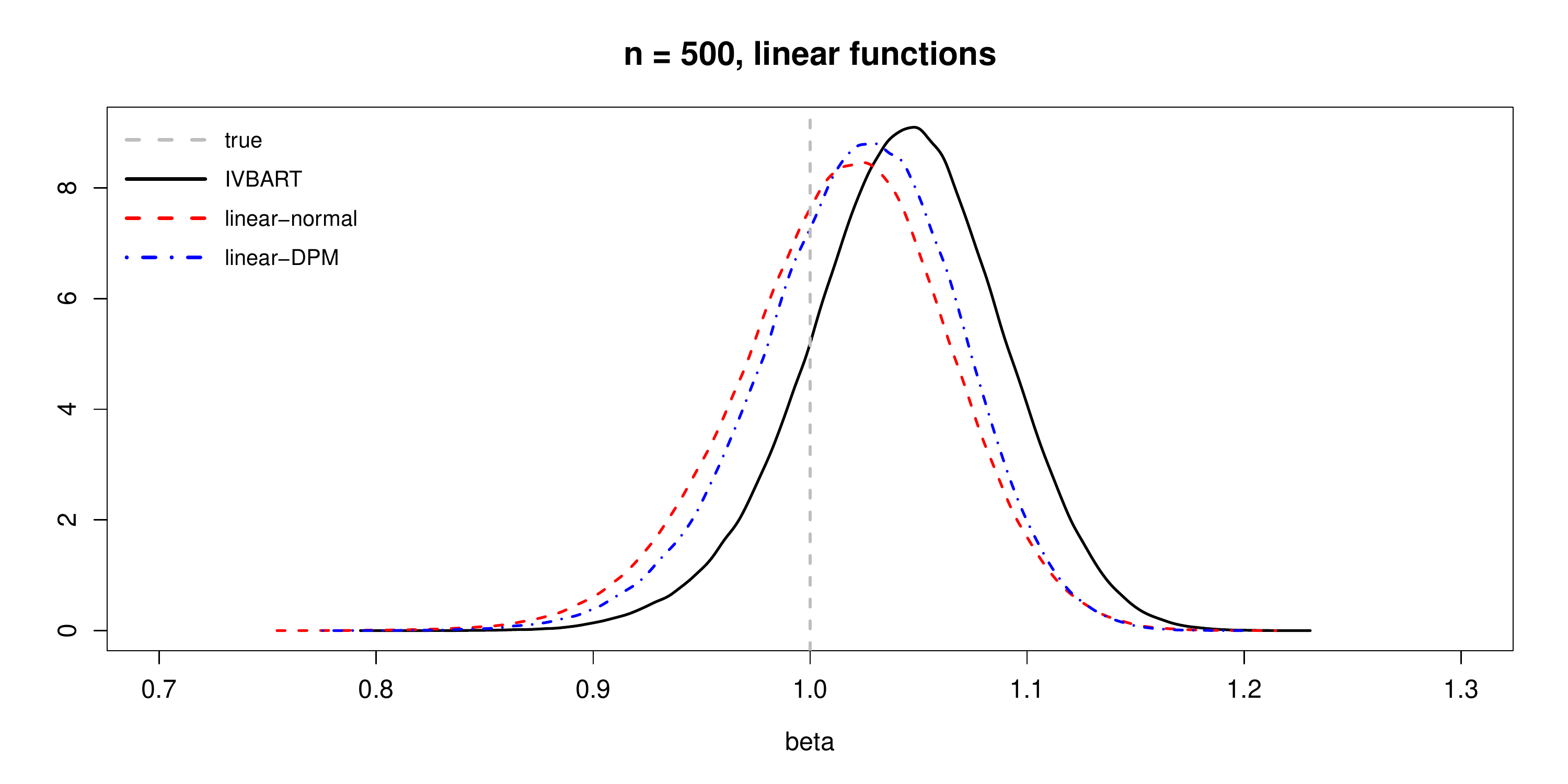} 
\caption{%
Densities estimates from MCMC draws of $\beta$ using the IVBART (solid), the linear-normal (dashed) and linear-DPM (dot-dash) models.
Each density estimate is based on all MCMC draws from all 90 data simulations.
In the top two figures, $n=2,000$.
In the bottom two figures, $n=500$.
In the left two figures, the data sets were simulated using the nonlinear function.
In the right two figures, the data sets were simulated using the linear functions.
\label{fig:alldraws}}
\end{figure}

Figure~\ref{fig:alldraws} displays the MCMC draws of $\beta$ from the three models
IVBART, linear-normal, and linear-DPM.
The four plots in the figure correspond to our four simulation scenarios.

The top-left plot of Figure~\ref{fig:alldraws}  corresponds to the scenario where we have simulated 2,000 observations (90 times) using
the nonlinear specifications of $f$ and $h$.
From each simulated data set we obtain a set of MCMC draws of $\beta$ and we combine all the draws into one large
set of draws and then use a density estimate to represent the draws.
The point of the paper is clearly illustrated by the  fact that the distribution of draws
from the IVBART model (solid density curve) is much tigher around the true value of $\beta=1$ than
the densities for the linear-normal model (dashed) or the linear-DPM model (dot-dash).
By figuring out the functions $f$ and $h$ from the data, with no user input, IVBART is able to get
a more precise inference for $\beta$ than is obtained by simply assuming the functions are linear.

From the top-right plot in Figure~\ref{fig:alldraws} we see that with $n=2,000$ and linear functions,
the inference from the IVBART model is very similar to that obtained from the two linear models linear-normal and 
linear-DPM.
The IVBART model is slightly more upward biased.

The two bottom plots of Figure~\ref{fig:alldraws} show that when the sample size is smaller, as we expect, 
things are tougher for the flexible model.
IVBART still produces draws closer to $\beta$ in the nonlinear case but there is some downward bias.
In the linear case, the IVBART draws are again slightly upwardly biased but still quite similar to the linear methods.

Table~\ref{tab:rmse} summarizes the results by reporting the root mean squared error (RMSE) of
the $\beta$ draws, again averaged over all MCMC draws and all simulations.
We also report the relative RMSE for each simulation scenario by dividing the RMSE of each model
by the minimum over the three models.
With $n = 2,000$ and nonlinear functions, IVBART has the smallest RMSE and the RMSE for the linear-normal model
is 86\% larger while the RMSE for the linear-DPM model is 77\% larger.
With $n = 500$, and nonlinear functions, IVBART is again the best with the linear-normal and linear-DPM
models being 42\% and 35\% worse.
In the linear cases, the linear models win, but the IVBART model is at most 28\% worse.
The numbers in Table~\ref{tab:rmse} reinforce the message of Figure~\ref{fig:alldraws}.
When there is strong nonlinearity, IVBART is much better and not too much worse in the linear case.

\begin{table}[tbp]  \centering
\begin{tabular}{l|rrr}
 & IVBART & linear-normal & linear-DPM \\
\hline
n=2,000, nonlinear & (0.022, 1.000) & (0.040, 1.858) & (0.038, 1.769) \\
n=2000, linear & (0.029, 1.275) & (0.024, 1.059) & (0.023, 1.000) \\
n=500, nonlinear & (0.060, 1.000) & (0.085, 1.417) & (0.080, 1.348) \\
n=500, linear & (0.062, 1.220) & (0.051, 1.000) & (0.051, 1.002)
\end{tabular}
\caption{%
RMSE and relative RMSE. 
Rows are for our four simulation scenarios and columns are for our three models.
Each table entry reports (RMSE, relative RMSE).
The relative RMSE for each simulation scenario is obtained by dividing the RMSE
for each of the three models by the minimum over the three models. 
\label{tab:rmse}
}
\end{table}

\begin{figure}
\centerline{\includegraphics[scale=.35]{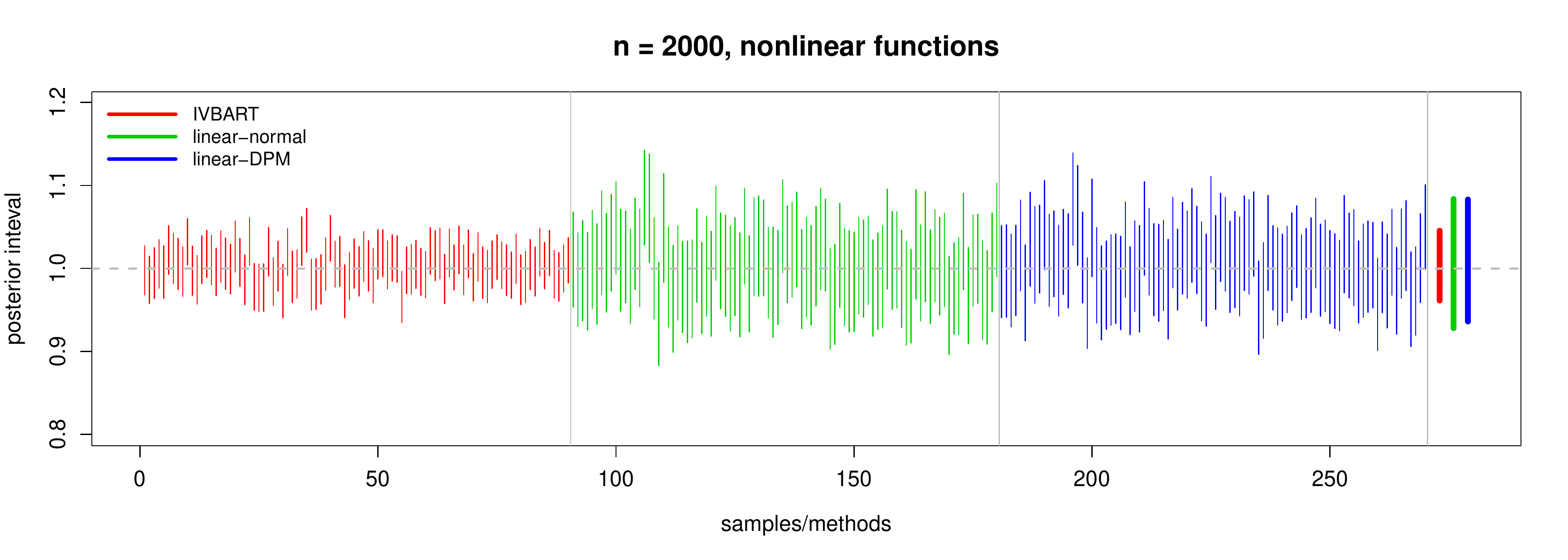}} 
\centerline{\includegraphics[scale=.35]{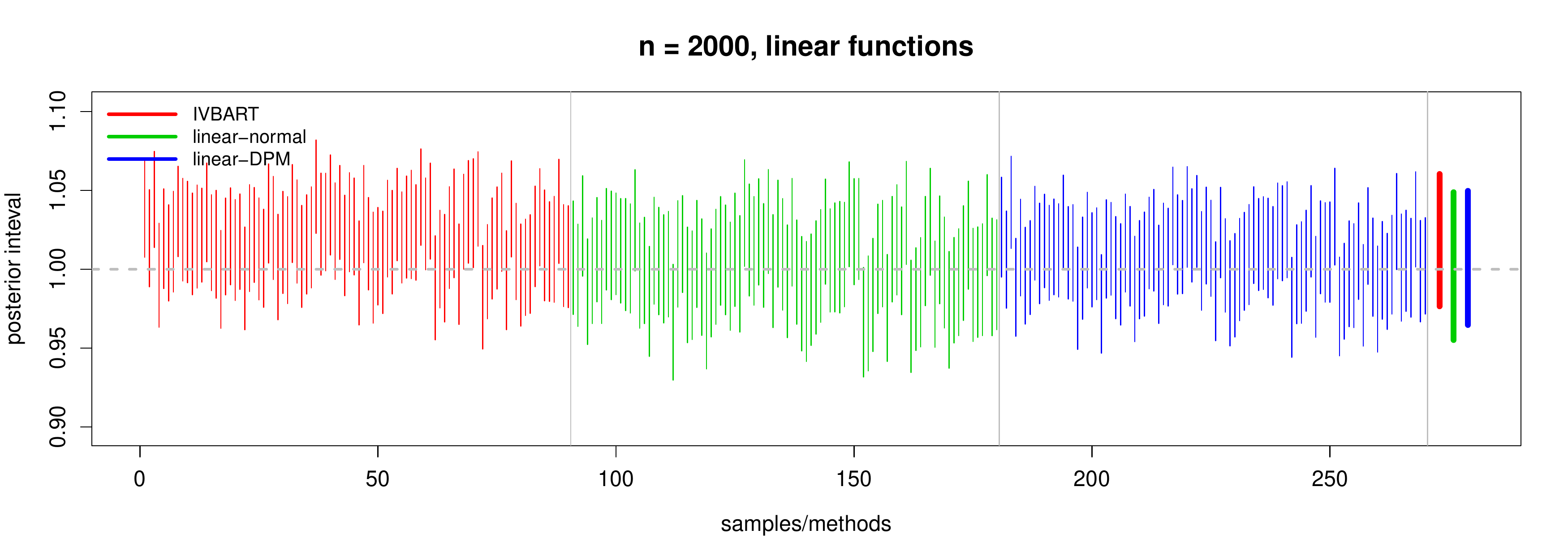}} 
\centerline{\includegraphics[scale=.35]{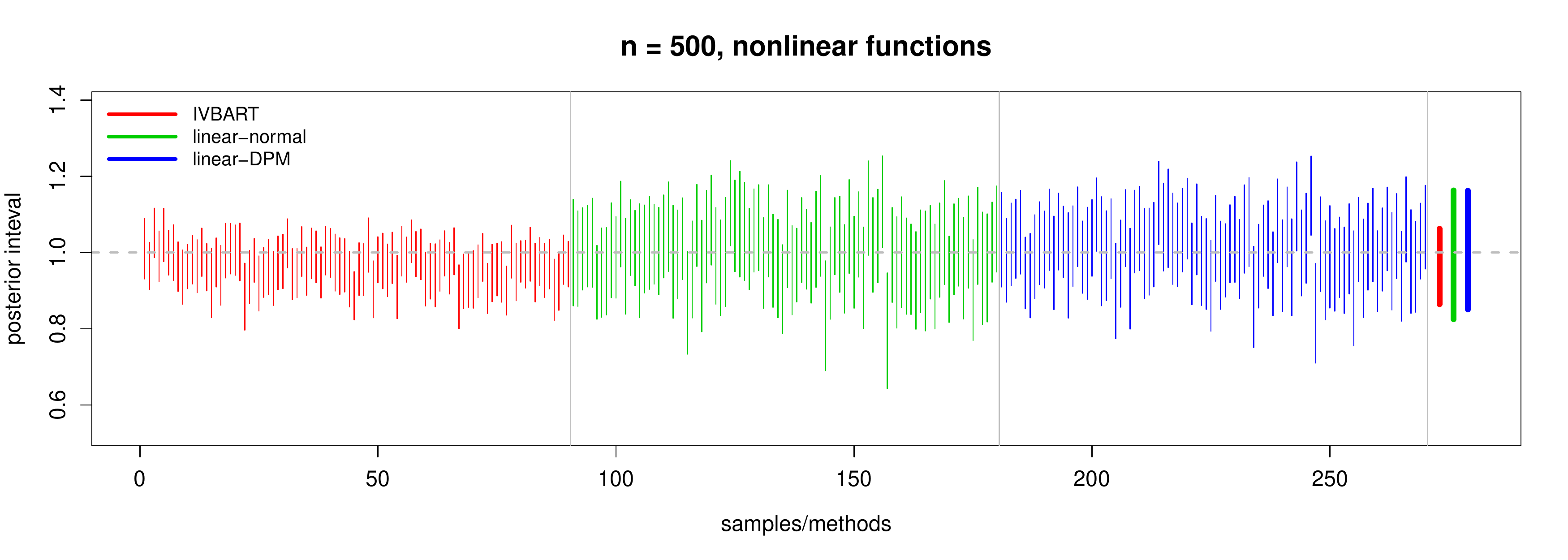}}  
\centerline{\includegraphics[scale=.35]{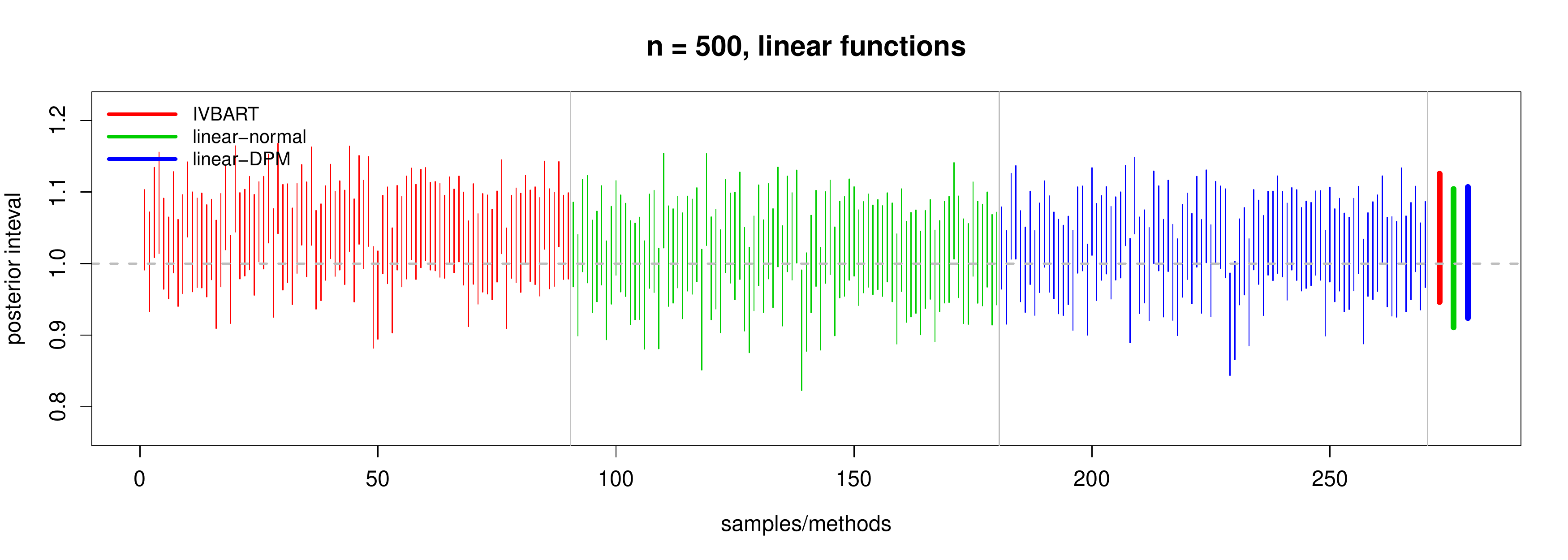}}
\caption{%
95\% Posterior intervals for $\beta$.
Each vertical segment represents a 95\% posterior interval.
Top two plots are for simlulation data sample size $n=2,000$ and the bottom two are for $n=500$.
Within each pair, the first is for the nonlinear functions and the second is for the linear functions.
Within each plot, the first 90 intervals are from IVBART while the second is from linear-normal and the third set of 90 is from linear-DPM.
The final three (thicker) intervals display .025 and.975 quantiles where all draw from all simulations are combined (as in Figure~\ref{fig:alldraws}).
\label{fig:sim-ints}}
\end{figure}

Figure~\ref{fig:sim-ints} displays 95\% posterior intervals for each simulation and each model.
The four plots again correspond to our four simulation scenarios.
Within each plot, each short vertical line segment represents a 95\% interval obtained from the 
.025 and .975 quantiles of the MCMC $\beta$ draws.
The first 90 line segments display the posterior intervals for the IVBART method while the 
second and thirds sets of 90 display the intervals for the linear-normal and linear-DPM models.
In each plot the final three (thicker) vertical line-seqments display the .025 and .975 quantiles for all draws
combined for each method (as in Figure~\ref{fig:alldraws} and Table~\ref{tab:rmse}).
In the top plot we clearly see the good performance of the IVBART model as the intervals are shorter
and located near the true value of $\beta$.
In the linear cases (second and fourth plot) we see that IVBART is not too different from the linear
models but somewhat biased upward.
In the nonlinear case with $n = 500$, the IVBART intervals are smaller than the linear ones but slightly
downard biased.
We discuss the bias further in Section~\ref{subsec:sim-prisens}.

\subsection{Prior Sensitivity}\label{subsec:sim-prisens}

In the bottom-left plot if Figure~\ref{fig:alldraws} we see a bias in the inference for $\beta$.
We also see this in the third plot in Figure~\ref{fig:sim-ints}.
Two basic features of our model may be contributing to this bias.
First, even when $f$ and $h$ are linear, our model is intrinsically nonlinear in inferring $\beta$.
This is made clear in the Gibbs conditional for $\beta$ given in Section~\ref{betacond}.
Secondly, the extreme flexibility of our model makes our inference sensitive to the prior.
In Equations \ref{n-liniv1} and \ref{n-liniv2}, both the nonlinear functions ($f$ or $h$) and
the error terms ($\epsilon_T$ or $\epsilon_Y$) are capable of adaptively capturing the variation on
$T$ and $Y$.
Of course, the degree to which the variation in $T$ and $Y$ is captured by the functions as opposed to
the errors, will affect our inference for $\beta$.
When the data are sufficiently informative (top-left of Figure~\ref{fig:alldraws}) the prior is less
influential.  But for smaller sample sizes (bottom-left of Figure~\ref{fig:alldraws}) the prior may affect
our inference.

In \citep{ChipGeor10} great care is taken to develop a data dependent prior for the error term
and nonlinear function for the simple single equation predictive model.
In \citep{DPMBART}, the approach is extended to a single equation with nonparametric error estimation.
While we are working to extend these approaches to our IV model, we first take the alternative approach
of studying the prior sensitivity.
In our current model, emphasis is on the estimation of the causal parameter $\beta$
as opposed the predictive goal emphasized in \citep{ChipGeor10}.
In this case, we find the prior sensitivity approach helpful.
See also \citep{BCF} for  an important contribution to the problem of model and prior specification when
using BART type models for causal inference.

The key prior choices involve the BART priors for $f$ and $h$ and the priors for the DPM estimation
of the joint error distribution of $(\epsilon_T,\epsilon_Y)$.
Details for these prior choices are given in Section~\ref{details}.
In this section we give an overview of the prior choices and examine the sensitivity of our
inference to a key aspect of the prior.

The prior for the error term estimation follows \citep{CHMR08,ROSSI14}.
We first rescale both $T$ and $Y$ by subtracting off the sample mean and then
dividing by the sample standard deviation.  Draws of $\beta$ are then rescaled to return to the original units.
The error DPM prior is then designed to be informative, but flexible enough to cover the full range of the data.
Of course the scaling based on the sample mean and standard deviation is sensitive to the error distribution
but \citep{CHMR08} report good results for severly non-normal errors.
We also note that the results reported in Section~\ref{subsec:sim-beta-inf} provide further evidence
for the excellent performance of the \citep{CHMR08} approach.
Note that for a single equation, this is a much more spread out prior for the errors than used
in \citep{ChipGeor10} or \citep{DPMBART}.

For the priors on $f$ and $h$ we start with the very simple BART prior specification:
\begin{equation}
f(x,z) \sim N(0,\sigma^2_f), \;\; h(x) \sim N(0,\sigma^2_h),
\end{equation}
where $\sigma_f$ and $\sigma_h$ are prior parameters that must be chosen.
This remarkably simple prior specification is key the success of BART.
Given we have standardize both $T$ and $Y$ simple prior choices could be
$\sigma_f \approx 1.0$ and $\sigma_h \approx 1$.
The default used for all results in Section~\ref{subsec:sim-beta-inf} are 
$\sigma_f = 1.2$ and $\sigma_h = 1.2$.

While these choices are simple and motived by the data standardization, they
may be too spread out in that both the error and the functions are allowed to capture
all of the variation.  In \citep{ChipGeor10} and \citep{DPMBART} the priors on the error process
are tuned to guide the model towards exploring inferences where the error is smaller.
We now explore the sensitivity of our results to the choices of $\sigma_f$ and $\sigma_h$.

\begin{figure}
\includegraphics[scale=.6]{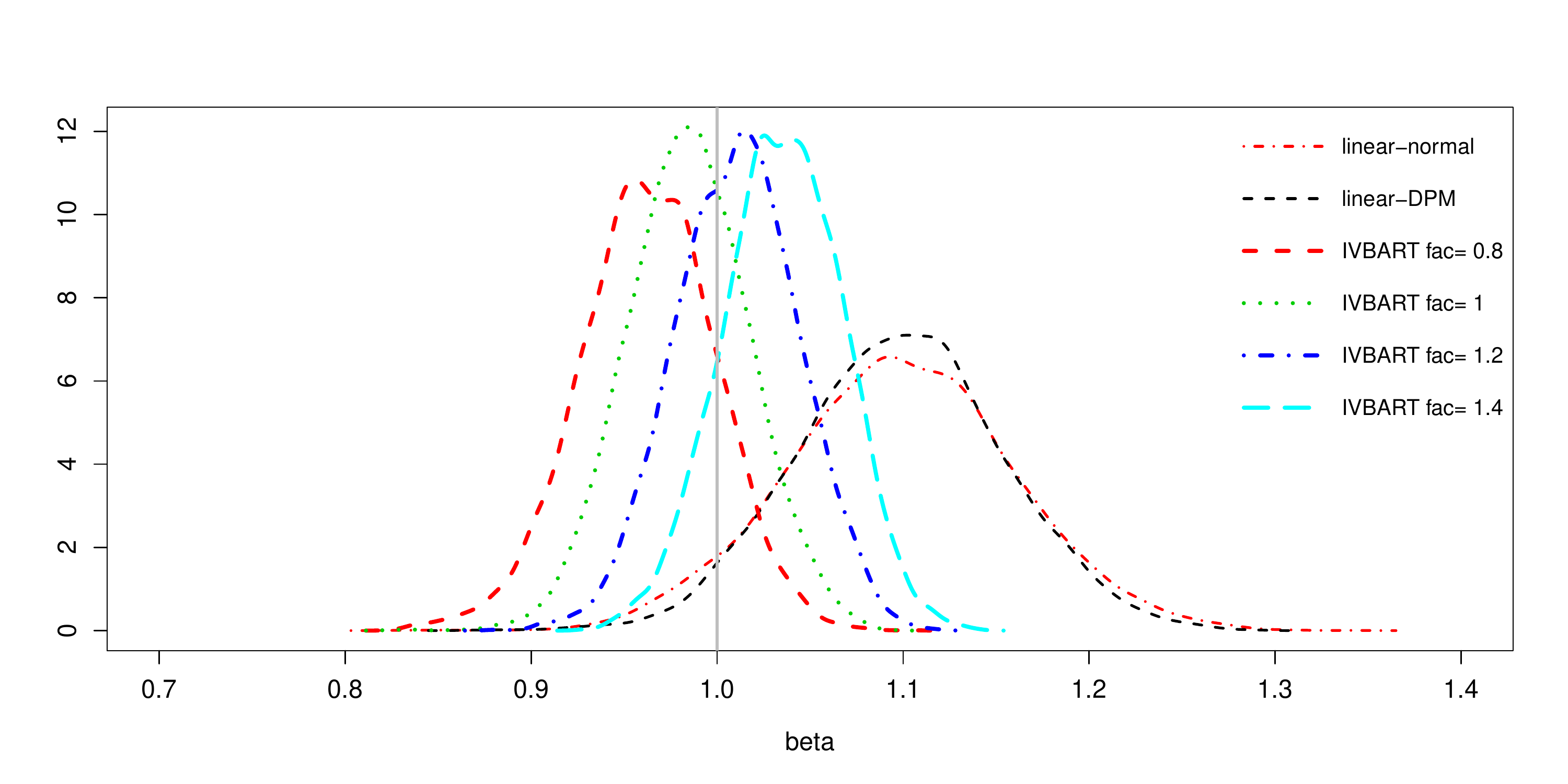} \\
\includegraphics[scale=.6]{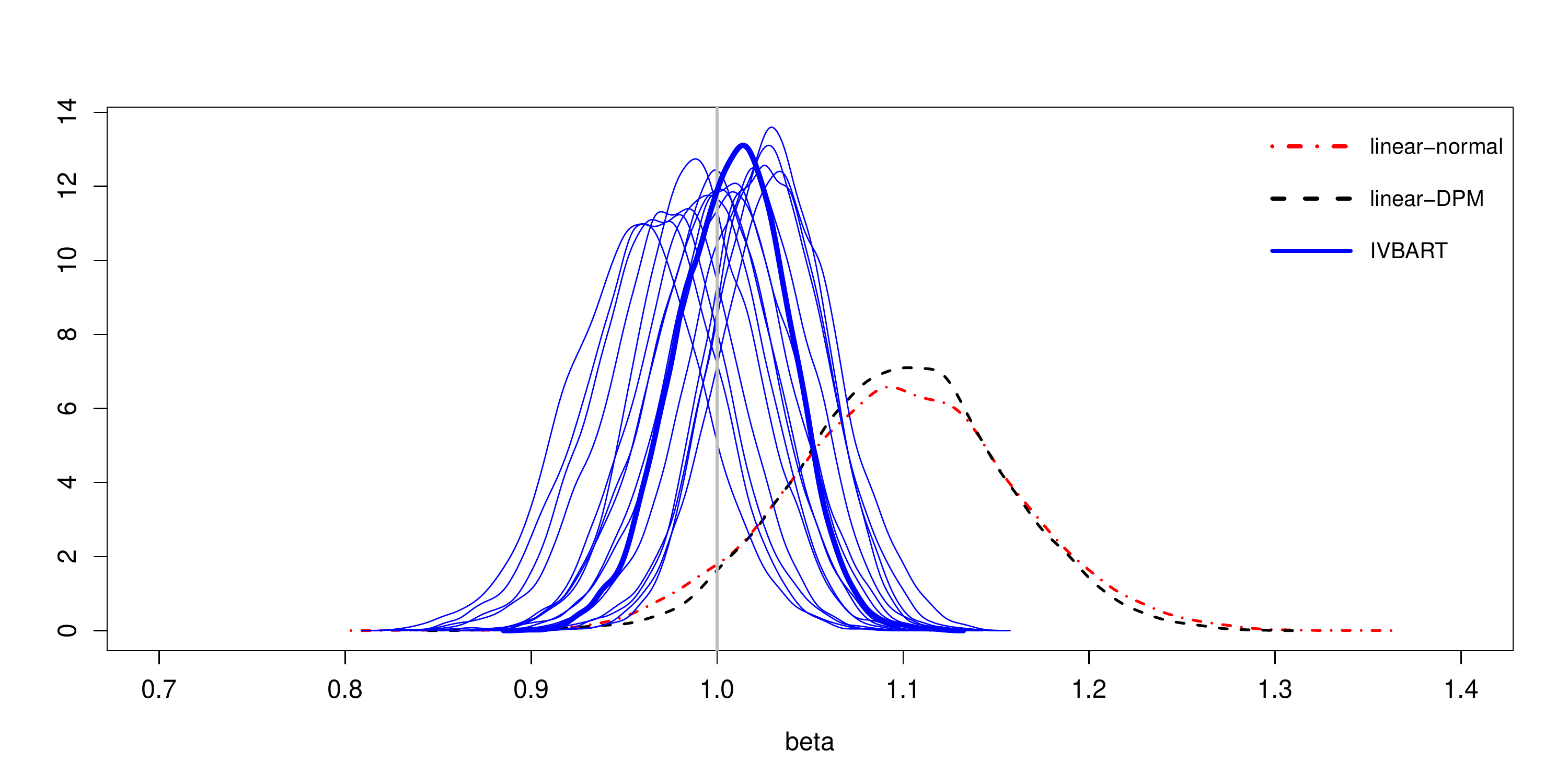} 
\caption{%
Prior sensitivity with $n = 500$.
In the top plot $\sigma_f = \sigma_h$ and these values are varied in $S_\sigma = \{.8,1,1.2,1.4\}$.
In the bottom plot, all 16 density estimates obtained using $\sigma_f \in S_\sigma$ and $\sigma_h \in S_\sigma$
are shown.
The thicker density corresponds to $(\sigma_f,\sigma_h) = (1.2, 1.2)$.
Densities for draws from the linear-normal (dot-dash line) and linear-DPM (dashed line) models are also shown in each plot.
\label{fig:sim-prisens-500}}
\end{figure}

\begin{figure}
\includegraphics[scale=.6]{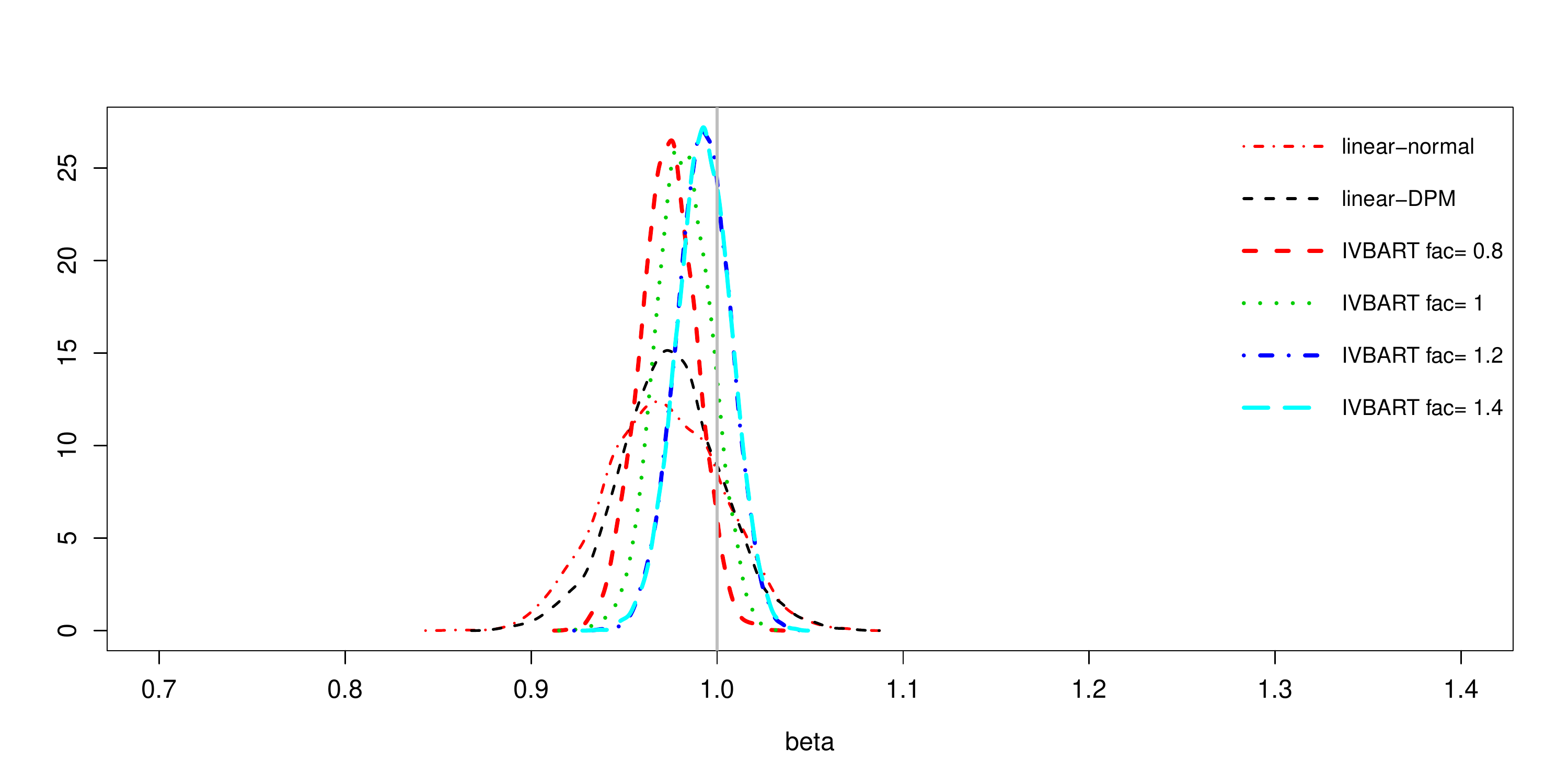} \\
\includegraphics[scale=.6]{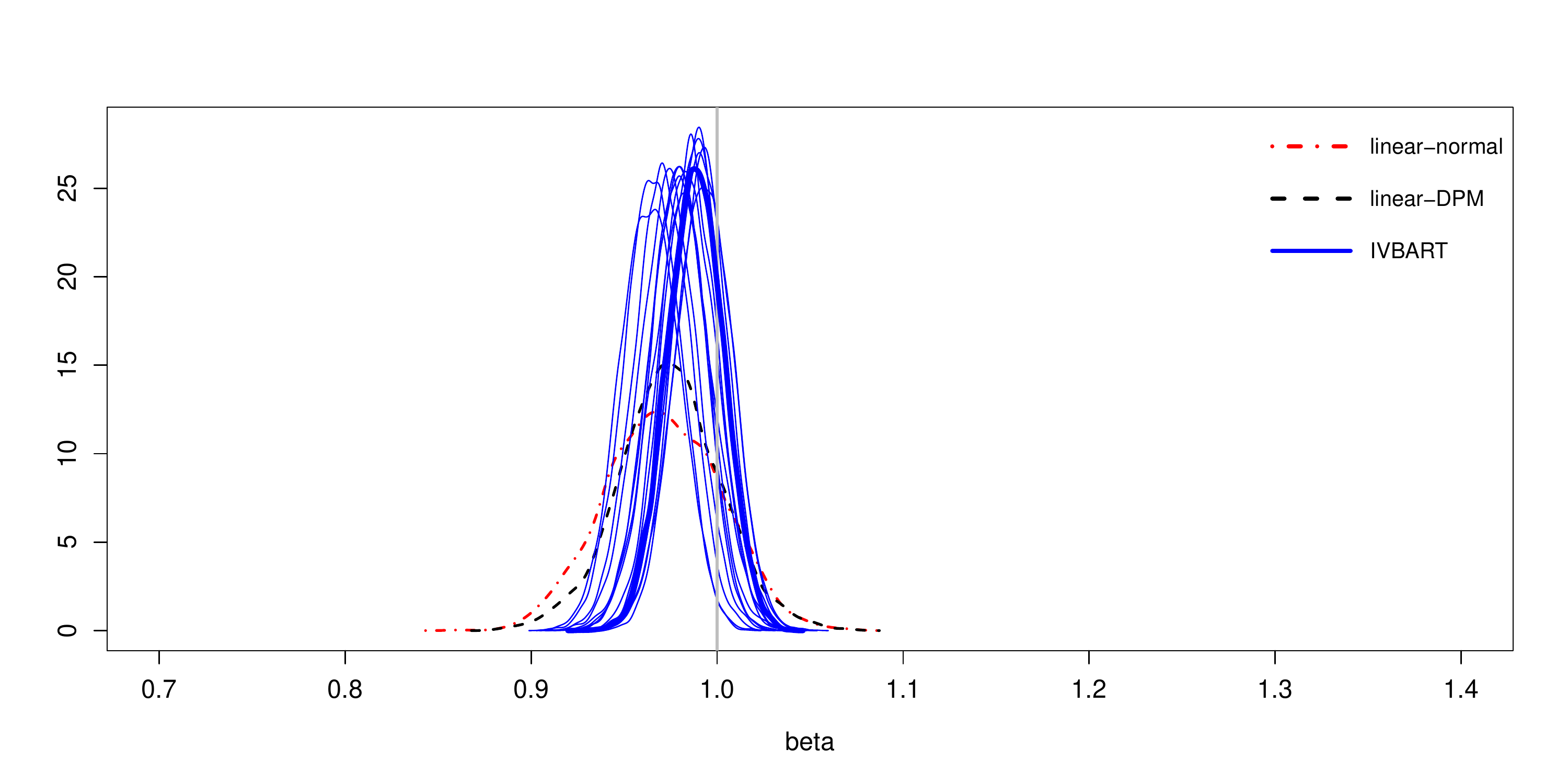} 
\caption{%
Prior sensitivity with $n = 2,000$.
In the top plot $\sigma_f = \sigma_h$ and these values are varied in $S_\sigma = \{.8,1,1.2,1.4\}$.
In the bottom plot, all 16 density estimates obtained using $\sigma_f \in S_\sigma$ and $\sigma_h \in S_\sigma$
are shown.
The thicker density corresponds to $(\sigma_f,\sigma_h) = (1.2, 1.2)$.
Densities for draws from the linear-normal (dot-dash line) and linear-DPM (dashed line) models are also shown in each plot.
\label{fig:sim-prisens-2000}}
\end{figure}

Figures \ref{fig:sim-prisens-500} and \ref{fig:sim-prisens-2000} present inference
for $\beta$ based on a single simulated data set.
Density estimates from MCMC draws of $\beta$ are presented where the prior choice is varied.
In Figure~\ref{fig:sim-prisens-500}, $n = 500$ while in Figure~\ref{fig:sim-prisens-2000}, $n = 2,000$.
In the top plot of each Figure, $\sigma_f$ and $\sigma_h$ are equal and varied in the set of values
$S_\sigma = \{.8,1,1.2,1.4\}$.
In the bottom plot of each figure all 16 possible combinations 
$\{(\sigma_f,\sigma_h): \sigma_f \in S_\sigma, \sigma_h \in S_\sigma \}$ are tried.
The thicker density  corresponds to the choice $(\sigma_f,\sigma_h) = (1.2, 1.2)$ used throughout
Section~\ref{subsec:sim-beta-inf}.

Clearly when is $n$ large (Figure~\ref{fig:sim-prisens-2000}) the results are fairly insensitive to the choice
of prior and indicative of a larger value for $\beta$ than suggested by the linear models.
When $n$ is smaller, (Figure~\ref{fig:sim-prisens-500}) the results are more sensitive to the prior, 
but we still have the correct suggestion that 
$\beta$ may be smaller than the values suggested by the linear models.

In practice we view the above sensitivity to be key part of the analysis as in Section~\ref{card} where
we analyze the famous Card data.
We are currenlty researching effective data based default prior choices, but feel than in this model
analysis of prior sensitivity will continue to be an essential part of the investigation.
Note that this is still much simpler than attempting to explore the sensitivity of two-stage least
squares to the inclusion of possible transformed $x$ and $z$.
As currently engineered, our approach is not targeted towards a ``big p'' scenario where we entertain
verly large $x$ or $z$ vectors of variables.
However, we feel the case we have investigated in our simulations with ten $x$ and five $z$ instruments 
is representative of many applied problems.
We explore the ``big p'' problem in future research.

\subsection{Markov Chain Monte Carlo Performance}\label{subsec:sim-mcmc}

\begin{figure}
\includegraphics[scale=.6]{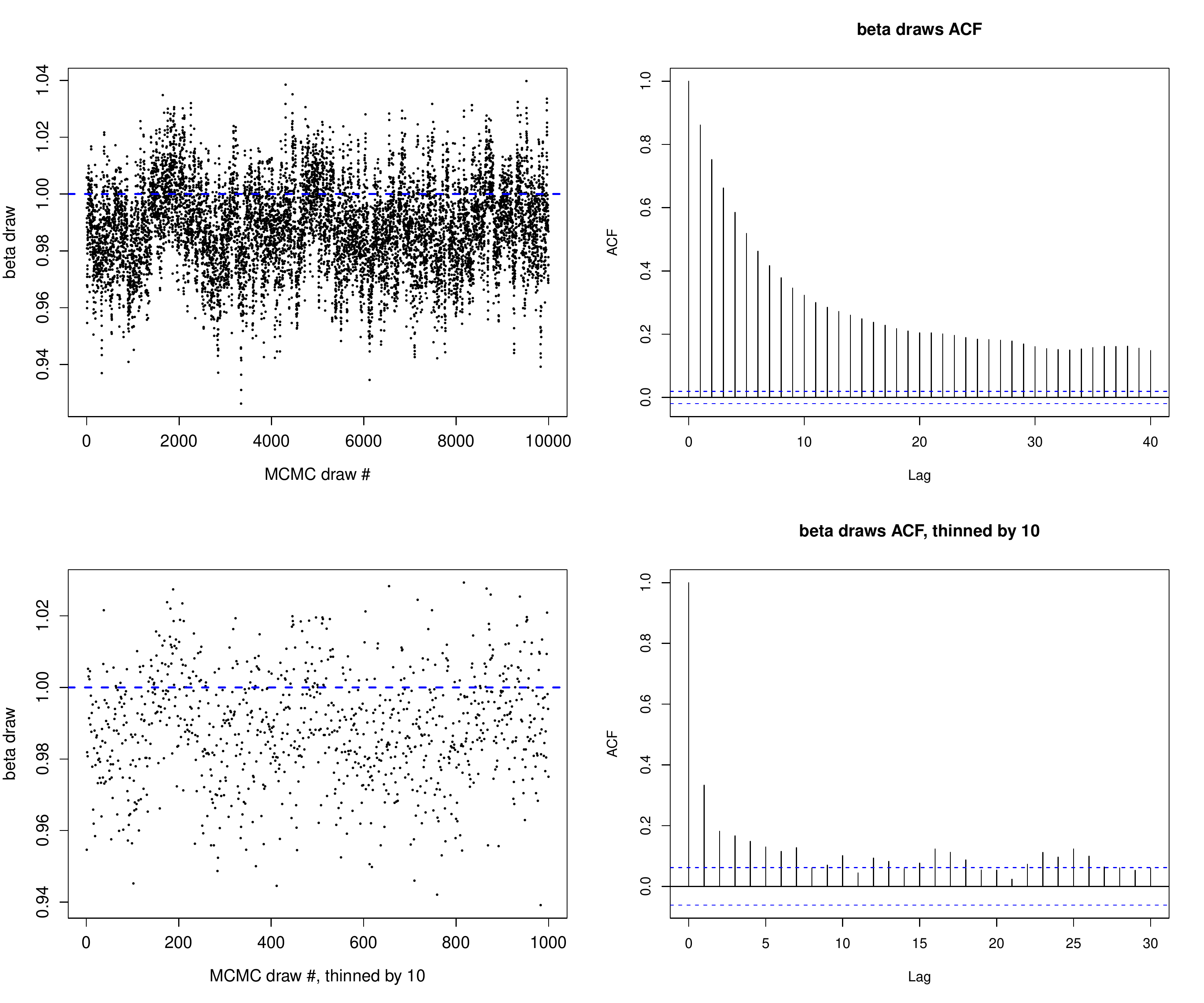}
\caption{%
Time series behaviour of the $\beta$ MCMC draws.
}\label{fig:mcmc}
\end{figure}

In Figure~\ref{fig:mcmc} we take a quick look at the time series characteristics of our
MCMC draws of $\beta$.
Figure~\ref{fig:mcmc} displays time series plots of the $\beta$ draws for a single drawn
sample of $n = 2,000$ observations and the nonlinear choices of $f$ and $h$.

The top left plot displays all draws and the top right displays the corresponding ACF.
The bottom left plot displays draws thinned to keep every tenth and the bottom right displays the corresponding ACF.

While the dependence is strong, we can obtain an effective inference in this case by simply taking
every thenth draw.

\section{Card Example}\label{card}

In a famous work \citep{Card93}, instrumental variables are used to
estimate the returns to education.  A standard specification of the
first stage regression relates the treatment variable
years-of-schooling by 1976 (\texttt{ed76} $=T$)  to two instruments
that measure how close a subject lives to a two- or a four-year
college ((\texttt{nearc2}, \texttt{nearc4}) $= z$)
and the confounders ($x$): 
years of experience by 1976 (\texttt{exp76}), 
years of experience squared (\texttt{exp762}), 
an African-American race indicator (\texttt{black}), 
an indicator for whether the subject lives in a
standard metropolitan statistical area in 1976 (\texttt{smsa76r}), and
an indicator for whether the subject lives in the south (\texttt{reg76r}).  
The measures of proximity to college are the
instruments in that they plausibly induce exogenous variation in the
cost of education and hence the amount of education.  The second stage
equation relates wages to the years of schooling and $x$.

Note that when running IVBART, we do not include \texttt{exp762} since the whole point of the model
is that the BART models for $f$ and $h$ are supposed to be able to uncover such nonlinearities without user input.
When running linear-normal and linear-DPM we do include \texttt{exp762}.

\begin{figure}
\centerline{\includegraphics[scale=.6]{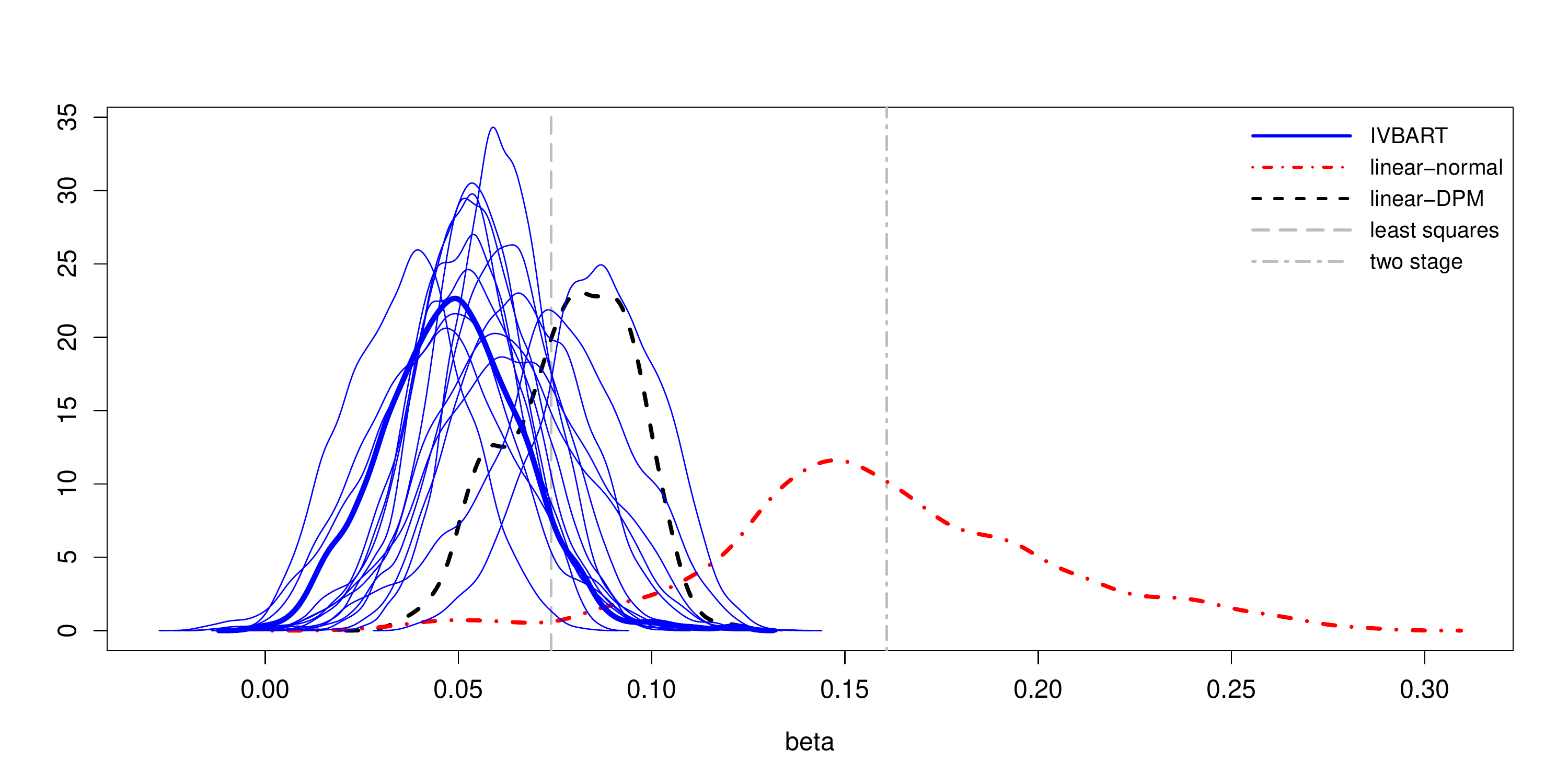}}
\caption{%
Inference for $\beta$ for the Card data.
To assess prior sensitivity, we vary both $\sigma_f$ and $\sigma_g$ 
in (.8,1.0,1.2,1.4) giving 16 possible choices for the pair.
All 16 IVBART posteriors are drawn with a solid curve.
The thicker IVBART line is for the setting $(\sigma_f,\sigma_h) = (1.2,1.2)$.
Densities
for draws from the linear-normal (dot-dash line) and linear-DPM (dashed line) models are
also shown.
\label{fig:card-data-2w-sens-1}}
\end{figure}

\begin{figure}
\centerline{\includegraphics[scale=.5]{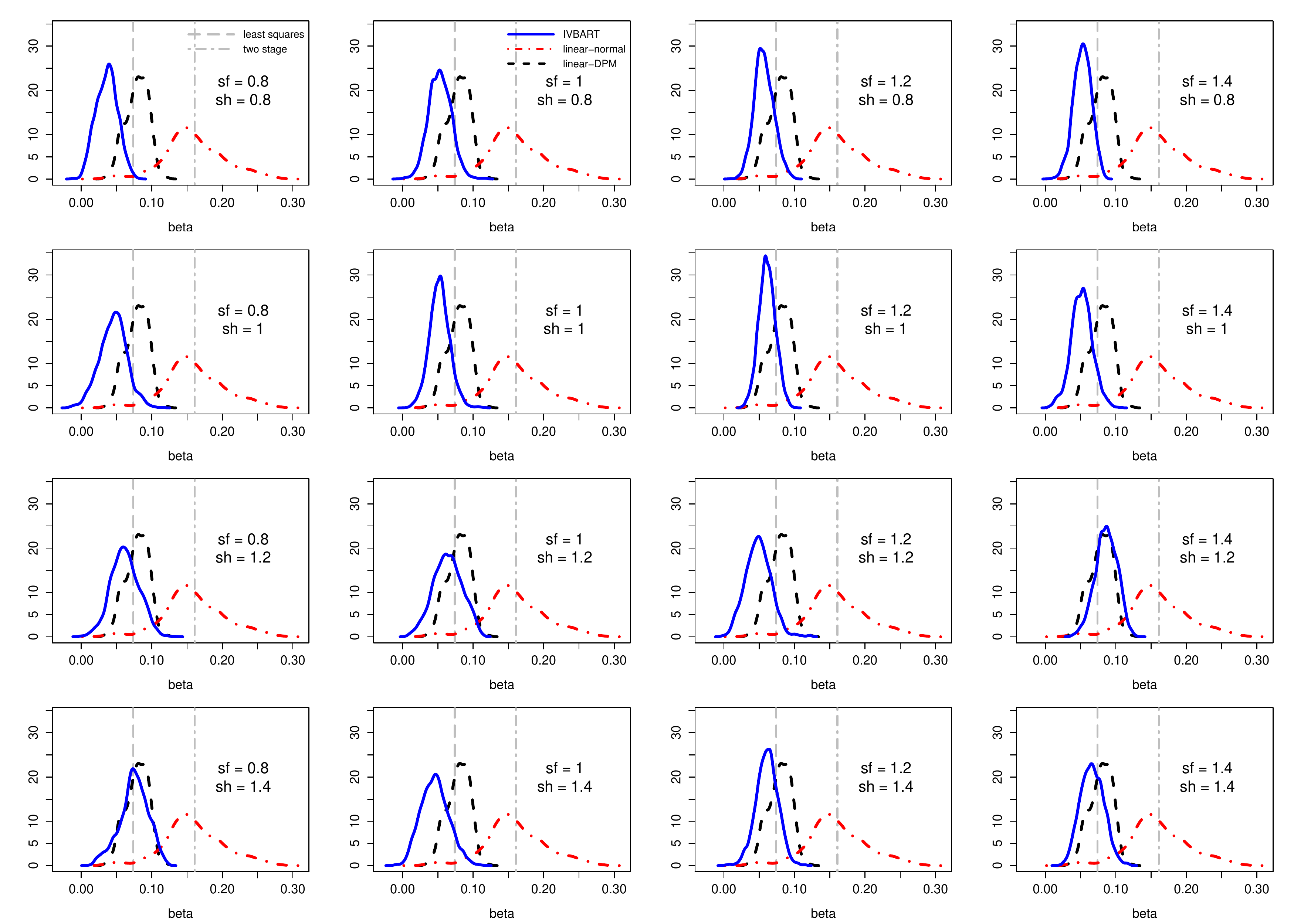}}
\caption{%
Inference for $\beta$ from the Card data.
To assess prior sensitivity, we vary both $\sigma_f$ and $\sigma_g$ in (.8,1.0,1.2,1.4) giving 16 possible choices for the pair.
Each plot in the figure corresponds to a different choice of $(\sigma_f,\sigma_h)$.
Densities for draws from the linear-normal (dot-dash line) and linear-DPM (dashed line) models are
also shown in each plot.
\label{fig:card-data-2w-sens-4}}
\end{figure}

Figures \ref{fig:card-data-2w-sens-1} and \ref{fig:card-data-2w-sens-4} display the inference
for $\beta$ using our three models IVBART, normal-linear, and normal-DPM.
The format of Figure~\ref{fig:card-data-2w-sens-1} is the same as that of 
the bottom plot in Figures \ref{fig:sim-prisens-500} and \ref{fig:sim-prisens-2000}
in which 16 IVBART inferences are displayed to capture the prior sensitivity to varying
both $\sigma_f$ and $\sigma_h$ in (.8,1.0,1.2,1.4).
For each of the 16 IVBART runs the posterior density of $\beta$ is displayed using a  solid curve.
The linear-model inference is displayed using a dash-dot curve and the linear-DPM inference is displayed
using a dashed curve.

The inference for $\beta$ from the linear-DPM (posterior mean .08)  
model suggests a value dramatically less than that suggested
by the linear-normal model (posterior mean .16).  
As expected, the posterior mean from  the linear-normal model is close to the estimate
obtained from standard two-stage least-squares (vertical dash-dot line).
Clearly, the IVBART inference suggests that, for reasonable priors, the value of $\beta$ may be less
than than suggested by the linear-DPM model.  However, the IVBART analysis still strongly supports the
belief that $\beta$ is far from zero as a practical matter with values around .05 being strongly favored.

Figure~\ref{fig:card-data-2w-sens-4} has the same information as Figure~\ref{fig:card-data-2w-sens-1}
but the density estimates of the posterior distribution of $\beta$ obtained from the 16 prior choices
are laid out in 16 separate plots.
There is one prior choice ($\sigma_f = 1.4$, $\sigma_h = 1.2$) such that the IVBART inference is very similar
to the linear-DPM inference.  However, for most choices that inference suggests a smaller value.
All IVBART posteriors suggest a value of $\beta$ much larger than zero.


Using IVBART we have obtained very strong inferences about the returns to schooling without having
to make any judgements about the fundamental functions $f$ and $h$.
This is much easier then searching through some catalogue of possible transformations however this is done.
Of course we still have the assumption of an additive linear treatment effect and relaxing this investigation
is a subject of our current research.
The sensitivity of the inference to the choice of the prior is an issue, but this is a natural consequence
of the flexibility of the model and the level of information in the data.

To quickly get a rough sense of the practical difference in the inferences show in Figure~\ref{fig:card-data-2w-sens-1},
we can say that according to the linear-normal, linear-DPM, and IVBART models, $\beta$ could be about .15, .08, or .05.
A change of 4 more years of schooling would then change y=log wage by .6, .32, and .2.
If we exponentiate these amounts, we get 1.8, 1.38, and 1.22 for the ratio of the wage level with and without the four years schooling.
All of these amounts are quite different from one as a practical matter and a 38\% increase in wages is quite a bit more than a 22\% increase.

\section{Details for the Gibbs Sampler}\label{gibbs}
In this section we provide some details for the Gibbs sampler.

For some of the development it will be useful to work in terms of the Cholesky root of 
$\Sigma_i$.

Let,
\begin{equation*}
L_i = \left[\begin{array}{cc}
\label{eq:L}
 \sigma_{Ti}  &  0\\
 \gamma_i & \sigma_{Yi}\\
 \end{array}\right]
\end{equation*}

so, that

$$
\Sigma_i = L_i \, L_i'.
$$

We can then write our model as:

\begin{eqnarray}\label{nliniv}
T_i & = & \mu_{Ti} + f(z_i,x_i) + \sigma_{Ti} \, Z_{Ti} \label{Liv1} \\
Y_i & = & \mu_{Yi} + \beta \, T_i + h(x_i) + \gamma_i Z_{Ti} + \sigma_{Yi} \, Z_{Yi} \label{Liv2}
\end{eqnarray}
where we recall that $\Sigma_i$ and the corresponding $(\sigma_{Ti}, \gamma_i, \sigma_{Yi})$ along with 
$(\mu_{Ti}, \mu_{Yi})$,  comprise the $\theta_i$ of Section~\ref{flex-mod}.

We now detail the four conditionals in the Gibbs sampler of Section~\ref{flex-mod}.
We present them in an order which we believe corresponds to increasing difficulty.
The first three are quite easy, while the last one, the draw of $f$, takes a little work.

Note also that for the model
\begin{equation}\label{bartpost}
Y_i = f(x_i) + \epsilon_i, \; \epsilon_i \sim N(0,w_i^2),
\end{equation}
with known $w_i$,
the BART prior and MCMC algorithm allows us to iterate a Markov Chain whose stationary distribution
the posterior of $f$.
We will have BART draws for both $f$ and $h$ conditional on the other parameters.
In each case we will see that we can write the information in the data in the form of 
Equation~\ref{bartpost} where the $Y_i$ and the $w_i$ depend on the data and the values of the
known parameters.

\subsection{The $\beta$ Conditional}\label{betacond}

Given all of the parameters except $\beta$ we can compute
\begin{equation}\label{zeq}
Z_{Ti} = (T_i - \mu_{Ti} + f(z_i,x_i))/\sigma_{Ti}
\end{equation}
from Equation~\ref{Liv1}.

We let
$$
V_i = (Y_i - \mu_{Yi} - h(x_i) - \gamma_i Z_{Ti})/\sigma_{Yi} \;\;  \mbox{and} \;\; W_i = T_i/\sigma_{Yi},
$$
from Equation~\ref{Liv2}.

This gives,
$$
V_i = \beta \, W_i + Z_{Yi}, \;\; Z_{Yi} 
\stackrel{\scriptstyle iid}{\sim} N(0,1)\ .
$$

Given the normal prior for $\beta$ we have standard normal draw for the conditional.

\subsection{The $h$ Conditional}\label{hcond}

This is similar to the $\beta$ conditional.

We then let
$$
V_i = (Y_i - \mu_{Yi} - \beta \, T_i - \gamma_i Z_{Ti}),
$$
from Equations \ref{zeq} and \ref{Liv2}.

This gives,
$$
V_i = h(x_i) + \sigma_{Yi} Z_{Yi},
$$
which allows for a BART draw of $h$ using \ref{bartpost}.

\subsection{The $\{\theta_i\}$ Conditional}\label{tcond}

Let,
\begin{equation}\label{tcond}
\tilde{Y} = (\tilde{Y}_{i1}, \tilde{Y}_{i2})' = (T_i - f(z_i,x), (Y_i - \beta \, T_i - h(x_i))'.
\end{equation}

Then,
$$
\tilde{Y}_i \sim N(\mu_i,\Sigma_i).
$$

Then, given $\{\tilde{Y}_i\}$, we can draw $\{\theta_i\} = \{(\mu_i,\Sigma_i)\}$ using the standard DPM 
methodology as described in \cite{ROSSI14}, \cite{CHMR08}, 
and originally in \cite{EW95}. 

\subsection{The $f$ Conditional}\label{fcond}

Finally, we draw $f$.

From Equations \ref{Liv1} and \ref{Liv2},

\begin{eqnarray*}
Y_i - \mu_{Yi} - h(x_i) & = & \beta \, T_i + \gamma_i \, Z_{Ti} + \sigma_{Yi} Z_{Yi} \\
                        & = & \beta (\mu_{Ti} +f(z_i,x_i) + \sigma_{Ti} Z_{Ti}) +  \gamma_i \, Z_{Ti} + \sigma_{Yi} Z_{Yi}.
\end{eqnarray*}

So,
$$
Y_i - \mu_{Yi} - h(x_i) - \beta \mu_{Ti} = \beta f(z_i,x_i) + Z_{Ti} (\beta \sigma_{Ti} + \gamma_i) + \sigma_{Yi} Z_{Yi}.
$$

We then let,
\begin{eqnarray*}
R_i & = & (\beta \sigma_{Ti} + \gamma_i)(T_i - \mu_{Ti}) - \sigma_{Ti}(Y_i - \mu_{Yi} - h(x_i) - \beta \mu_{Ti}) \\
    & = & \gamma_i f(z_i,x_i) - \sigma_{Ti} \sigma_{Yi} Z_{Yi}.
\end{eqnarray*}

Thus, for each $i=1,2,\ldots,n$ we have the pair of independent observations,
$$
T_i - \mu_{Ti} = f(z_i,x_i) +  \sigma_{Zi} Z_{Ti}, \;\; 
\frac{R_i}{\gamma_i} = f(z_i,x_i) - \frac{\sigma_{Ti} \sigma_{Yi}}{\gamma_i} Z_{Yi}.
$$

This gives us $2n$ observations which may be put in the form of Equation~\ref{bartpost}.

Note that if $|\gamma_i|$ is small, then we automatically throw out the information in the 
$\frac{R_i}{\gamma_i}$ observation since the resulting large error variance will downweight the observation.
This makes intuitive sense since if $|\gamma_i|$ is small the errors in the two equations are independent so
that our information about $f$ comes soley from the first equation.

\section{Prior Details}\label{details}
In this section we provide details on the choice of prior.
As illustrated in sections \ref{simulated} and \ref{card}, the choice of prior is influential.
This is inevitable in a flexible Bayesian model.
In our basic model (equations \ref{n-liniv1} and \ref{n-liniv2}) the nature of $(T,Y)$
can be captured by nonparametrically modeling the error $(\epsilon_T,\epsilon_Y)$ or the functions
$f$ and $h$ and we are doing both.

Priors must be chosen for $\beta$, the functions $f$ and $h$, and the Dirichlet process mixture for
$\epsilon$.  We discuss each of these in turn.

As in many Bayesian analyses, we want to be able to inject prior information when available and we want
reasonable defaults that enable users to get sensible results with minimal input.
In order to have sensible default choices we typically start by standardizing the data.
In all of the examples run in Sections \ref{simulated} and \ref{card} we  started by stardizing the data
to have zero mean and standard deviation one:
\begin{equation}\label{eq-scale}
T \rightarrow \frac{T - \bar{T}}{s_T}, \;\;  Y \rightarrow \frac{Y - \bar{Y}}{s_Y}
\end{equation}
where $\bar{x}$ and $s_x$ are the sample mean and standard deviation of the data in $x$.

\subsection{Prior on $\beta$}\label{subsec:beta-prior}

While the simple linear specification for the treatment effect is a limitation, it facilitates
the very simple prior specification
\begin{equation}\label{eq.beta-prior}
\beta \sim N(\bar{\beta},A_\beta^{-1}).
\end{equation}

Important prior information about $\beta$ may well be available in application.
As often the information in the data is not overwhelmingly strong, inclusion of sensible prior information
may be an essential part of the analysis.

Note that if we standardize the data as in (\ref{eq-scale}), 
then
\begin{equation*}
\beta_s = \beta \, \frac{s_T}{s_Y},
\end{equation*}
where $\beta_s$ is the coefficient on the standardized scale and $\beta$ is the coefficient on the original
scale.

In all the examples, the prior in (\ref{eq.beta-prior}) is applied to $\beta_s$.
The posterior draws of $\beta_s$ are then transformed back to the original $\beta$ scale.

\subsection{Priors for $f$ and $h$}\label{subsec:fh-prior}

A major strength of the BART approach is the remarkably simple specification for the prior on an unknown 
function.  We have:
\begin{equation}
f(z,x) \sim N(0,\sigma_f^2), \;\; h(x) \sim N(0,\sigma_h^2).
\end{equation}
We need only choose the two standard deviations $\sigma_f$ and $\sigma_h$.
Note that the marginal prior for $f(z,x)$ does not depend on $(z,x)$.  Similarly, the prior
for $h(x)$ does not depend on $x$.

There are additional details to the full BART specification.
For example, there are prior choices that describe beliefs about the trees underlying 
the functions $f$ and $h$.  All such choices are done as discussed in \citep{ChipGeor10} and
implemented in the package \citep{BARTRP} in \citep{RENV}.

Given the data has been standardized, a diffuse but hopefully not too spread out prior is obtained
by letting $\sigma_f$ and $\sigma_h$ be in the neighborhood of one.
In our examples, our exploration of prior sensitivity consists of varying $\sigma_f$ and $\sigma_f$ about one
with the choice $\sigma_f = \sigma_h = 1.2$ being highlighted 
(Figures \ref{fig:sim-prisens-500}, \ref{fig:sim-prisens-2000}, \ref{fig:card-data-2w-sens-1}, and \ref{fig:card-data-2w-sens-4}).
The motivation for the choice 1.2 is that it gives a prior which allows for more variation than the more obvious choice of 1, 
while hopefully not being too spread out.
More spread out priors (that is, larger $\sigma_f$ and $\sigma_h$) give $f$ and $h$ more freedom to fit the data.
Of course, we live in constant fear of over-fitting.
As is standard practice in applied Machine Learning we could reasonably use some kind of out-of-sample test
to guide our choices.

This approach to choosing $\sigma_f$ and $\sigma_h$ 
is roughly in accordance with the standard choice in \citep{ChipGeor10} and \citep{BARTRP}.
There, the default is chosen so that twice the standard deviation of $f$ covers the range of $Y$ in the simpler
model $Y = f(x) + \epsilon$.
However, even our simple two equation IV model is highly nonlinear and the consequences of prior choices
may be hard to anticipate.  These considerations motivate the prior sensitivity approach taken in 
Sections \ref{simulated} and \ref{card}.
We explore values of $\sigma_f$ and $\sigma_h$ in neighborhoods of one.

\subsection{Dirichlet process mixture Prior}\label{subsec:DPM-prior}

In this section we describe the choice of $G_0$ and prior on $\alpha$.
Recall (Section~\ref{flex-mod}) that the atoms of the discrete distribution from which we draw $\theta_i = (\mu_i,\Sigma_i)$ are draws
form $G_0$ and $\alpha$ determines the distribution of the number of unique $\theta_i$.

Our choices follow \citep{CHMR08} exactly.  
In particular we review the basic rational and argue that the same choices are reasonable in our more flexible model.
As previously noted, \citep{ROSSI14} is also an excellent reference, giving a less terse textbook style presentation.

The basic idea is to calibrate these fundamental prior choices by considering the scenario where $\beta$, $f$, and $h$ are all zero.
In this case, $(\epsilon_{Ti},\epsilon_{Yi})' = (T_i,Y_i)'$ and 
our DPM model should nonparametically estimate the 
bivariate joint distribution of the standardized $(T,Y)$.  
These choices are ``noninformative'' but not so spread out as to limit the effectiveness of the DPM.
The examples in \cite{CHMR08}  and Section~\ref{simulated}
suggest that these choices are quite generally effective in the linear case.

Note however that this approach deviates from the approach motivating the prior choices made in \citep{ChipGeor10}.
In \citep{ChipGeor10}, we have the single equation $Y = f(x) + \epsilon$ with $\epsilon \sim N(0,\sigma^2)$. 
The data on $Y$ are demeaned, $f$ is shrunk to zero, and the prior on $\sigma$ is chosen to suggest that $f$ will fit $Y$ better
than a linear function would.
That is, the prior on the single parameter $\sigma$ is designed to suggest that it is smaller than that obtained from
a linear fit.  Here the prior is calibrated be even more spread out that needed to completely fit the data.
The \cite{CHMR08} DPM prior is less informative about the errors so that our prior sensitivity approach is useful in uncovering
the range of plausible inferences.
Note that in \citep{DPMBART}, a single equation $Y = f(x) + \epsilon$ was considered and the DPM modeling approach was used
to univariate distribution of $\epsilon$.  In that paper, the DPM choices were motivated by a desire to mimic the kind of prior
information used in \citep{ChipGeor10} rather than the relatively noninformative specification used here for the joint distribution
of $(\epsilon_{T},\epsilon_{Y})'$.

\subsection{Specification of $G_0$}


The base prior $G_0$ is a prior on $\theta = (\mu,\Sigma)$.
We start from the standard conjugate setup:
$$
\Sigma^{-1} \sim \text{Wishart}_\nu(V^{-1}), \;\; \mu \,|\, \Sigma \sim N(\bar{\mu}, \frac{\Sigma}{a}).
$$
The parametrization of the Wishart distribution is such that $E(\Sigma^{-1}) = \nu \, V^{-1}$.

Given our standardization of $T$ and $Y$, $\bar{\mu}$ is set to zero.  We also let $V = v \, I$ where $I$ is the $2 \times 2$ identity matrix.

With these simplifications we only have to choose the three numbers $(a,\nu,v)$.
Using $\sigma_1 = \sqrt{\sigma_{11}}$,
we first choose $c_1,c_2,c_3,\kappa$ and then find $(a,\nu,v)$ such that
$$
P(-c_3 < \mu_1 < c_3) = 1 - \kappa, \;\; P(\sigma_1 < c_1) = \kappa/2, \, P(\sigma_1 > c_2) = \kappa/2,
$$
so that $P(c_1 < \sigma_1 < c_2) = 1-\kappa$.
The defaults used throughout this paper are $c_3 = 10$, $c_1 = .25$, $c_2 = 3.25$, and $\kappa = .2$,
giving $a = .016$, $\nu = 2.004$ and $v=.17$.
Again, this is exactly as in \citep{CHMR08}.

The value of $c_3$ is very large and the value of $\nu$ is very small.
These priors are chosen to be very diffuse but not so diffuse as to derail our basic DPM MCMC algorithm.


Note that the marginals from the conjugate prior are analytically available with,
$$
\sigma_{11} \sim \frac{v}{\chi^2_{\nu-1}}, \; \text{and} \; \mu_1 \sim \sqrt{\frac{v}{a(\nu-1)}} \, t_{\nu-1}.
$$

\subsection{Prior on $\alpha$}



The idea of the prior is to relate $\alpha$ to the number of unique $\theta_i$.
Let $I$ denote the number of unique $\theta_i$. 
The user chooses a minimum and maximum number of components $I_{min}$ and $I_{max}$.
We then solve for $\alpha_{min}$ so that the mode of the consequent distribution for $I$
is $I_{min}$.
Similarly, we obtain $\alpha_{max}$ from $I_{max}$.
We then let
$$
p(\alpha) \propto (1 - \frac{\alpha - \alpha_{min}}{\alpha_{max} - \alpha_{min}})^\psi.
$$
The default values for $I_{min}$, $I_{max}$, and $\psi$ are 2, $[.1n]+1$, and .5, where
$[]$ denotes the integer part  and $n$ is the sample size.
A nice thing about this prior is it automatically scales sensibly with $n$.

\section{Conclusion}\label{sec:conclusion}

The linear instrumental variables model has long been fundamental in causal analysis.
It simply and elegantly captures the fundamental intuition
that an instrumental variable $z$, may provide
a source of variation in a treatment $T$, comparable to that of an experiment
in which
variation is induced by an investigator who
controls the value of $T$. 

However the assumption of linearity is rarely one that we can  comfortably impose.
In practice, this usually leads to a search for a set of transformations of the instruments
$z$ and the additional variables $x$ which are then used in the linear setting.
Even with modern methods for finding transformations this process is tedious and depends on
choices for the set of transformations considered.

Our use of Bayesian Additive Regression trees (BART) allows us to capture a wide
range of possible functions with no user input and still do a full Bayesian analysis including nonparametric
modeling of the error terms.

For our nonparametric error term analysis we have followed \cite{CHMR08} closely given its success.
This as led to a prior-sensitivity approach in which we vary the prior beliefs about the nonlinear functions $f$ and $h$.
Also, our goal here is inferential in that we seek to learn $\beta$ while in BART, the prior development has been
more focused on the goal of out of sample prediction.
The BART models for these two functions allow for a relatively simple scheme for varying our prior beliefs.
We hope that the top-left plot of Figure~\ref{fig:alldraws} and the analysis of the Card data in Figure~\ref{fig:card-data-2w-sens-1}
will suggest to practictioners that IVBART provides a relatively simple alternative to the difficult challenges presented
by the general sensitivity of inference for the treatment effect $\beta$ to the model specification.

In future work we will consider the use of more informative data based priors for the error distribution 
as in \citep{ChipGeor10} and \citep{DPMBART}.
In addition, future work will seek to relax the additive linear assumption for the treatment effect.
While are current analysis is very flexible and allows for simple interpretation of the causal effect through
the parameter $\beta$, we wish to consider the possibility of hetergeneous treatment effects.
We note that our current model is already very flexible and powerful and extentions to a still more flexible model
will entail careful prior choices as in \citep{BCF}.

We note that the simlulation study presented in this paper provides further support for the
efficacy of the linear approaches provided by the \texttt{R} package
\texttt{bayesm} \citep{RPbayesm} in the functions
\texttt{rivGibbs} for the linear model with correlated normal errors and \texttt{rivDP} for the linear model
with nonparametrically modeled errors.



\section{Acknowledgment}

Research reported in this publication was supported in part by the
National Cancer Institute of the National Institutes of Health under
award number RC4CA155846.  The content is solely the responsibility of
the authors and does not necessarily represent the official views of
the National Institutes of Health.
 
\newpage
\appendix

\section{The ivbart R package}\label{ivbartR}

All of the calculations in this article with respect to the IVBART
model were performed with the {\bf ivbart R} package.  {\bf ivbart}
is free open-source software that is publicly available at
\url{https://github.com/rsparapa/bnptools}.  The following
snippet of {\bf R} code installs {\bf ivbart} with the 
{\tt install\_github}
function from the {\bf remotes R} package (available on
the Comprehensive R Archive Network at 
\url{https://cran.r-project.org/package=remotes}).
\begin{verbatim}
R> library("remotes")
R> install_github("rsparapa/bnptools/ivbart")
\end{verbatim}
The {\tt nlsym} object is a data frame providing the Card example data
of Section~\ref{card}.  With {\tt system.file("demo/nlsym.R",
  package="ivbart")}, you can find the installed {\bf R} program that
analyzes the Card data with IVBART that is provided 
as a demonstration.  You can run this program with the following
snippet.
\begin{verbatim}
R> source(system.file("demo/nlsym.R", package="ivbart"), echo=TRUE)
\end{verbatim}

See the documentation of the {\tt ivbart} function for more details.

\section{Causal Identification}\label{appendix}

In this section, we prove that the estimation of $\beta$ is causally
identified.  First, let's return to the structural equations in the
classic IV framework.  Here, we ignore the confounders for simplicity
since they are not needed, i.e., we can simply assume that they are
unobserved.  Furthermore, let the constant intercept terms be zero
for convenience, i.e., $\mu_T=\mu_Y=0$.  And, finally, we substitute
the first stage into the second stage.
\begin{align}
T_i & = \gamma' Z_i  + \epsilon_{Ti} \label{tsls1} \\
Y_i & = \beta \, T_i + \epsilon_{Yi} \nonumber \\
Y_i &=  \beta (\gamma' Z_i  + \epsilon_{Ti}) + \epsilon_{Yi} \label{tsls2}
\end{align}

We will show that $\beta$ is identifiable by resorting to the
so-called instrumental variable formula \citep{BowdTurk90}.
\begin{align}\label{IVformula}
\beta & \triangleq \pderivf{ \E{Y|T} }{T} \nonumber \\
& \triangleq \frac{\E{Y|Z}}{\E{T|Z}} \\
& \triangleq \frac{r_{YZ}}{r_{TZ}} \nonumber 
\end{align}
The middle formula \eqref{IVformula}, a ratio of expectations, is the
key to our proof of causal identification; rather than the last line
which is the most cited form of this result.
To apply the middle formula, we plug the first stage into
the denominator \eqref{tsls1} and the re-written second stage into the
numerator \eqref{tsls2} to show that $\beta$ is identifiable.
\begin{align*}
\beta & \triangleq \frac{\E{Y|Z}}{\E{T|Z}} \\
& = \frac{\beta \gamma' Z_i }{\gamma' Z_i } = \beta
\end{align*}
This is a well-known result with respect to linear structural equations.

Now, let's investigate our BART IV framework while, once again, ignoring
confounders and letting the intercepts be zero.
\begin{align}
T_i & =  f(Z_i) + \epsilon_{Ti} \label{ivbart1} \\
Y_i & =  \beta \, T_i + \epsilon_{Yi}  \nonumber \\
   & =  \beta (f(Z_i) + \epsilon_{Ti}) + \epsilon_{Yi} \label{ivbart2}
\end{align}
And, we apply the instrumental variable formula \eqref{IVformula} as
before to show that $\beta$ is identifiable, i.e., plug the first stage into
the denominator \eqref{ivbart1} and the re-written second stage into the
numerator \eqref{ivbart2}.
\begin{align*}
\beta & \triangleq \frac{\E{Y|Z}}{\E{T|Z}} \\
& = \frac{\beta f(Z_i) }{ f(Z_i) } = \beta
\end{align*}
This is a more surprising result.  Generally, it is well-known that
nonparametric methods are not identifiable without strong assumptions
\citep{ImbeAngr94,Pear09}.  We illustrate the typical non-idenfiability
of $\beta$ by a more general model as follows.
\begin{align}
T_i & =  f(Z_i) + \epsilon_{Ti} \label{nonpar1} \\
Y_i & =  g(T_i) + \epsilon_{Yi}  \nonumber \\
    & =  g(f(Z_i) + \epsilon_{Ti}) + \epsilon_{Yi} \label{nonpar2}
\end{align}
Now, apply the instrumental variable formula \eqref{IVformula}, i.e.,
plug the first stage into the denominator \eqref{nonpar1} and the
re-written second stage into the numerator \eqref{nonpar2}.
\begin{align*}
\frac{\E{Y|Z}}{\E{T|Z}} & = \frac{\E{g(f(Z_i) + \epsilon_{Ti})} }{ f(Z_i) } 
 \not= \beta
\end{align*}
The denominator is unchanged.  However, the numerator
does not have a simple form; therefore, the true
value $\beta$ is not identifiable without further assumptions
about $g(.)$.  For example, if we assume that $g(T_i)=\beta\, T_i$
(as we have above), then $\beta$ is identifiable as we have shown.

\newpage
\bibliographystyle{abbrvnat}
\bibliography{ivbart}

\end{document}